\pdfoutput=1
\documentclass[11pt]{article}

\usepackage{acl}

\usepackage{todonotes}
\usepackage{times}
\usepackage{latexsym}
\usepackage{graphicx}
\usepackage{multirow}
\usepackage{algorithm}
\usepackage{algorithmic}
\usepackage{booktabs} 
\usepackage{tabularx}
\usepackage{multirow}
\usepackage{amsmath}
\usepackage{adjustbox}
\usepackage{cleveref}
\usepackage{diagbox}
\usepackage{tabularx}
\usepackage{tcolorbox}
\tcbuselibrary{breakable,skins}
\usepackage{enumitem}
\usepackage{amsmath}
\usepackage{amssymb}

\usepackage[T1]{fontenc}
\usepackage[utf8]{inputenc}

\usepackage{microtype}

\usepackage{inconsolata}

\title{[Research-proposal] Mapping the Demographic Gradient: How Annotator Traits Shape LLM-Human Alignment}

\title{[Research-proposal] How Much is Too Much? Finding the Right Demographic Mix for Human Alignment in LLMs }

\title{The Cost of Demographic Over-Specification: LLM Agreement with Matched Human Judgments}

\title{Demographic Prompting at Scale: When More Attributes Hurt LLM--Human Agreement}

\author{
\textbf{Mahammed Kamruzzaman}, \textbf{Shrabon Das}, \textbf{Gene Louis Kim} \\
$^{1}$University of South Florida, $^{2}$North South University \\
$^{1}$\{kamruzzaman1, almonsur, das157, genekim\}@usf.edu, $^{2}$enamul.hassan@northsouth.edu
}

\author{
\textbf{Mahammed Kamruzzaman}, \textbf{Shrabon Kumar Das}, \textbf{Gene Louis Kim} \\
Bellini College of AI, Cybersecurity and Computing\\
University of South Florida \\
\{kamruzzaman1, das157, genekim,\}@usf.edu
}

\begin{document}

\maketitle

\begin{abstract}
We investigate how annotator demographic attributes, supplied as prompt cues, shape the alignment between large language model (LLM) predictions and human annotations across five tasks. Using five open-source LLMs, we systematically vary the number and composition of demographic components in the prompt, spanning every combination from single-attribute through full-attribute configurations. Our experiments reveal three principal findings. First, alignment consistently peaks with one to three high-signal attributes and degrades under the full attribute set, establishing a clear over-specification threshold. Second, the overall magnitude of demographic influence on human annotations does not predict which attributes improve LLM alignment; instead, both the learnability and the directional coherence of each attribute's annotation signal need to be considered jointly. Third, neuron probing reveals that specialized activation correlates with alignment gains only under coherent annotation signals, and that activation volume alone does not imply steerability. Together, these results demonstrate that demographic prompting is not a monolithic intervention: its utility is highly context-dependent, shaped by attribute signal quality, task characteristics, and model architecture.
\end{abstract}

\section{Introduction}

LLMs are increasingly used as substitutes for, or complements to, human annotators on subjective NLP tasks such as toxicity detection, sentiment analysis, and offensiveness rating \cite{gilardi2023chatgpt, tornberg2023chatgpt}. Because these tasks reflect annotator subjectivity, a natural question arises: how well can an LLM actually model a given demographic perspective when prompted to do so? A growing body of work shows that LLM outputs are not demographically neutral, predictions tend to align more closely with certain demographic groups in the absence of demographic cues, and explicitly incorporating such cues into the prompt can shift model behaviour in ways that are neither uniform nor always beneficial \cite{beck-etal-2024-sensitivity,sun-etal-2025-sociodemographic,alipour-etal-2025-robustness,kamruzzaman-etal-2025-anger,schafer-etal-2025-demographics}.

Despite this progress, the existing literature leaves several important questions unresolved. Most prior studies examine only one or two specific demographic attributes at a time, or compare a no-demographic baseline against a single all-attributes prompt, without exploring the combinatorial space in between \cite{beck-etal-2024-sensitivity,sun-etal-2025-sociodemographic,alipour-etal-2025-robustness}. None systematically chart the incremental trajectory from single-attribute through multi-attribute to full-attribute prompting, %and none ask whether there is an optimal number of demographic components beyond which additional information introduces noise rather than signal. 
nor investigate the signal to noise relationship along this trajectory. 
To address this, we pose our first research question: \textbf{RQ1:} \emph{To what extent do individual versus combined annotator demographic attributes shape LLM--human alignment, which demographic features (or combinations) most significantly affect alignment, and is there a threshold beyond which additional demographic information no longer benefits alignment? 
%is no longer helpful?
%provides diminishing returns?
}

While several studies report that demographic prompting sometimes helps and sometimes hurts alignment \cite{gupta2023bias,brown2025evaluating,kamruzzaman2024woman}, the field lacks a principled account of \emph{why}: what properties of a demographic attribute, at the dataset level, predict whether prompting with it will improve a given model's agreement with human annotators? An attribute may strongly predict variation in human labels yet carry internally opposed subgroup signals that no single persona prompt can resolve. Understanding this requires moving beyond aggregate importance measures to characterise the structural quality of each attribute's annotation signal. We therefore ask: \textbf{RQ2:} \emph{To what extent do dataset-level demographic signals predict when demographic prompting improves LLM--human alignment, and do LLM outputs reflect the learnability and directional coherence of those signals?}

Finally, the internal mechanisms by which LLMs process demographic cues remain poorly understood. Neuron-level interpretability methods have been applied to multilingual and cultural knowledge \cite{ying-etal-2025-disentangling, zhao2023survey}, but no prior work has connected neuron activation patterns to demographic alignment in subjective language understanding annotation tasks. If a demographic cue activates specialized neurons within the model, does that internal engagement translate to better alignment with the prompted group's annotation norms, or can high activation volume coexist with poor steerability? This motivates our third research question: \textbf{RQ3:} \emph{To what extent do specialized neuron activations explain the variations in LLM--human alignment across demographic configurations and tasks?}

To answer these questions, we conduct experiments across five subjective language understanding task 
% \gknote{I want to avoid using ``sociolinguistic'' for reasons we discussed on Friday. We should replace all instances of ``sociolinguistic dataset'' with something like ``social judgment dataset'', ``subjective language understanding task'', or ``affective evaluation task'' instead. I'll let you choose the exact phrase you'd like to use.} 
datasets, each annotated with demographic metadata, on five open-source LLMs. Overall, our work makes the following contributions:
 
\begin{enumerate}[nosep,leftmargin=*]
    \item \textbf{Full combinatorial analysis of demographic prompting.} To the best of our knowledge, we are the first to systematically enumerate every combination of demographic attributes, from single-component through full-attribute configurations, across five tasks and five models. This reveals a consistent pattern: alignment peaks at one to three attributes and degrades under the full set, a finding that prior work, limited to single-attribute or all-attributes comparisons, could not establish.
 
    \item \textbf{A three-level framework linking dataset-side signal to LLM alignment.} We show that the magnitude of demographic influence on human annotations is uninformative for predicting alignment, while the learnability of word-demographic interaction patterns is a better but still insufficient predictor. The missing piece is directional coherence: attributes whose subgroups agree on which words signal the task label enable alignment gains, while attributes with opposed subgroup signals invert the relationship between learnability and alignment. This three-level characterisation (magnitude, learnability, coherence) provides a principled diagnostic to decide which attributes to include in demographic prompts.
 
    \item \textbf{First application of neuron probing to demographic alignment.} By adapting the specialised-neuron methodology of \citet{ying-etal-2025-disentangling} to the demographic prompting setting, we show that neuron activation proportion correlates with alignment only when the underlying annotation signal is directionally coherent. We also document the DeepSeek high-volume paradox, where the model activating the most specialised neurons is the least demographically steerable, demonstrating that activation quantity does not dictate
    %is dissociable from 
    alignment utility.
\end{enumerate}

\section{Related Work}

\paragraph{Sociodemographic prompting.}
The idea of conditioning LLM predictions on annotator demographics has been explored in several recent studies, though each addresses only a subset of the design space we consider. \citet{beck-etal-2024-sensitivity} examine five attributes across seven datasets and find that single-attribute prompts alter predictions while the all-attributes prompt causes the most label changes, but no intermediate combinations are tested. \citet{sun-etal-2025-sociodemographic} find that models align most with White annotators on politeness and offensiveness, and that demographic cues rarely improve alignment. Similarly, \citet{schafer-etal-2025-demographics} report that default model behaviour already leans toward White and younger viewpoints. \citet{alipour-etal-2025-robustness} show that confounders such as document difficulty and annotator sensitivity explain more variance than demographics alone, using logistic regression to model alignment as a function of both which is methodologically closest to our RQ2, though we focus on the structural properties of the demographic signal itself. 
% More broadly, LLMs tend to align more with female annotators \citep{de2025effects}, yet model choice and prompting strategy remain stronger predictors of alignment than demographic variables \citep{brown2025evaluating}. 
Personas can also trigger biases, mirroring gender-emotion stereotypes \citep{plaza-del-arco-etal-2024-angry} and producing regional disparities in emotion attribution \citep{kamruzzaman-etal-2025-anger}.

\paragraph{Dataset annotator demographic analysis.}
A complementary line of research examines how annotator demographics shape the annotation data itself, independent of LLMs. \citet{hovy-etal-2013-learning} present methods to model annotator reliability, while \citet{davani2022dealing} argue that aggregation can suppress minority viewpoints. \citet{orlikowski2023ecological} demonstrate that individual annotator differences often exceed demographic-group differences, cautioning against ecological fallacy in demographic alignment claims. Our SHAP and LinearSVC analyses (\Cref{sec:directional_coherence}) operate at the dataset level, %but we connect these dataset-side quantities 
which are connected 
to LLM behavior through rank correlation and Fisher-averaged coherence, bridging the gap between annotation analysis and prompting outcomes.

\paragraph{Mechanistic interpretability.}
Our neuron probing approach builds on work linking neuron activations to knowledge use in LLMs. Prior studies identify language-selective neurons that improve multilingual accuracy~\cite{tang-etal-2024-language,zhao2024large} and show that activations encode recoverable parametric knowledge~\cite{hong-etal-2025-intrinsic}. Most closely aligned to our paper, \citet{ying-etal-2025-disentangling} show that activating more specialized neurons corresponds to better cultural knowledge use, while \citet{cao2025model} argue that activation quantity does not necessarily reflect knowledge quality. Internal activations have also been used to steer model behavior~\cite{rimsky-etal-2024-steering} and mitigate bias through neuron editing~\cite{liu2024devil,yu2025understanding}. Unlike these approaches, we use neuron activations purely as a diagnostic signal to explain when demographic prompting succeeds or fails.

\section{Experimental Setup}

\subsection{Datasets}
We use five datasets where annotator demographic information is recorded alongside their task-specific annotations.

\begin{enumerate}[leftmargin=*,nosep]
    \item \textbf{Toxicity Detection:} For toxicity detection, we use the \textit{Diverse Perspectives (DP)} dataset \cite{kumar2021designing}. We use eight annotator demographic attributes for this dataset: \textit{age\_range, education, gender, is\_parent, lgbtq\_status, political\_affiliation, race, and religion\_importance.}

    \item \textbf{Sentiment Analysis:} \citeposs{diaz2018addressing} sentiment dataset studies age-related bias in sentiment analysis. The nine demographic attributes in this dataset are: \textit{age, education, employment\_status, gender, hispanic\_latino, income, living\_situation, political\_id, and race.}

    \item \textbf{Politeness:} \citeposs{pei-jurgens-2023-annotator} politeness dataset is a subset of the POPQUORN dataset \cite{pei-jurgens-2023-annotator}, with five annotator demographic attributes: \textit{age, education, gender, occupation, and race.}

    \item \textbf{Offensiveness:} We use the offensiveness subset of the POPQUORN dataset. This dataset has the same five demographic annotator attributes as the politeness task.

    \item \textbf{Emotion Attribution:} We use the International Survey on Emotion Antecedents and Reactions (ISEAR) \cite{scherer1994evidence}, which has seven demographic attributes: \textit{gender, religion, father's\_occupation, mother's\_occupation, field\_of\_study, country, and age.}
\end{enumerate}
% We use all demographic attributes in each respective dataset without modification. See \Cref{app:detail_dataset} for dataset details.
All of these demographic attributes are originally included in their respective datasets; we did not create or modify them. See \Cref{app:detail_dataset} for dataset details.
%For example, the politeness dataset already includes five annotator demographic attributes. For more details about these datasets, see \Cref{app:detail_dataset}.

\subsection{LLMs} We use five open-source LLMs: LLama-3.2-3B, Mistral-7B, Gemma3-12B, Qwen2.5-7B, and DeepSeek-R1-7B. For model details, see \Cref{app:llm_detail}.

\subsection{Prompts}  We prompt the LLMs both with demographic attributes and without (baseline). Prompts follow a common structure, with task-specific variation reflecting the nature of the task and available annotation labels:
%For the five tasks in our study, we use five different prompts, meaning the prompts are task-dependent, although they follow a common general structure. All task-specific prompts are provided in Appendix C. 
%A general prompt structure with demographic attributes is:
\textit{``Given the following text: \{text\}, how would a person of \{demographic\_attributes\} rate this \{task\}?''}. \{demographic\_attributes\} can be instantiated with any combination of demographic attributes available in the dataset. 
%in the following ways: \textit{Single-Demographic Prompting}, % (e.g., only race or only gender), 
%\textit{Multi-Demographic Prompting} (combinations of 2, 3, ... N demographic attributes), and \textit{All-Demographics Prompting} (using the full demographic metadata available for each sample). 
The \{task\} component is adjusted depending on the specific task being evaluated. All task-specific prompts are provided in \Cref{app:detail_prompt}.
To ensure the robustness of our findings, experiments include two additional paraphrases of the prompting template 
%we also use two additional rephrased versions of the prompting template 
and report the average results across all %three
prompting templates.

\subsection{Model-human Alignment Metrics} We quantify model–human alignment\footnote{Ongoing alignment research debates whether models should represent a broad ``average user'' or multiple demographic groups~\cite{schwerzmann2025desired,korinek2022aligned,hristova2024problem}. We \textbf{do not} investigate which alignment scale is normatively preferable; rather, we ask: when demographic attributes are provided in the prompt, to what extent do LLM outputs actually reflect the associated annotation patterns?} with quadratic-weighted \textit{Cohen's}~$\kappa$~\cite{cohen1968weighted}, a chance-corrected agreement coefficient for ordered rating scales. The quadratic penalty assigns partial credit to near misses and treats distant disagreements as total mismatches, capturing the ordinal structure of our five-point toxicity, sentiment, politeness, and offensiveness labels while allowing direct comparison with human–human reliability standards~\cite{landis1977measurement}. Cohen's $\kappa$ ranges from $-1$ to $+1$, where $+1$ indicates perfect agreement, $0$ agreement no better than chance, and $-1$ systematic disagreement. We report 95\% bootstrap confidence intervals and additionally compute \textit{macro} and \textit{micro} accuracy. \textbf{\textit{For the Emotion dataset only, we omit weighted $\kappa$ as the labels in this dataset have no natural ordinality.}} We use weighted $\kappa$ as the primary metric for discussing the results, as it largely correlates with macro accuracy.

\subsection{Dataset-Side Demographic Signal Analysis}
\label{sec:dataset_singal_Setup}

To quantify the demographic signal present in each dataset, we employ \textbf{three} complementary analyses that characterize each attribute's signal at increasing levels of structural detail. \textbf{First}, to measure the \textbf{\emph{magnitude}} of each attribute's influence on human label variation, we train a Logistic Regression classifier on one-hot-encoded annotator demographic attributes and apply SHAP (SHapley Additive exPlanations)~\cite{lundberg2017unified}, aggregating absolute SHAP values of binary indicator features back to their parent demographic groups (e.g., combining \texttt{gender\_Female} and \texttt{gender\_Male} into a single \texttt{gender} importance score). This analysis is conducted entirely on the annotation data; no LLM outputs are involved. Full details are in \Cref{app:shap}.
% \textbf{Second}, to assess the \textbf{\emph{learnability}} of each attribute's lexical signal, we train a LinearSVC (LSVC) with explicit \textit{word$\times$demographic interaction} features (computed as the element-wise product $X_{\text{inter},j} = X_{\text{words}} \odot d_j$, where $X_{\text{words}}$ is the TF-IDF matrix of text features and $d_j$ is the demographic indicator vector for category $j$). 
\textbf{Second}, to assess the \textbf{\emph{learnability}} of each attribute's lexical signal, we train a LinearSVC (LSVC) with explicit \textit{word$\times$demographic interaction} features, computed as the element-wise product:
\begin{equation}
X_{\text{inter},j} = X_{\text{words}} \odot d_j
\label{eq:interaction}
\end{equation}
where $X_{\text{words}}$ is the TF-IDF matrix of text features and $d_j$ is the demographic indicator vector for category $j$.
This yields a separate learned weight for every (word, demographic category) pair, so the resulting LSVC $\kappa$ reflects how much discriminative lexical signal each attribute contributes beyond text alone.
\textbf{Third}, to characterize the \textbf{\emph{directional coherence}} of each attribute's signal, we extract the top-200 most influential words per attribute from the fitted LSVC interaction weights 
and compute Spearman rank correlations between the weight vectors of every pair of demographic categories (i.e., subgroups) within each attribute (e.g., Female vs.\ Male within \texttt{gender}, or Liberal vs.\ Conservative within \texttt{political\_affiliation}), summarized via the Fisher $z$-transformation into a single Fisher-averaged $\bar{\rho}$ per attribute. A positive $\bar{\rho}$ means subgroups weight the same words similarly (a coherent signal a prompted LLM could in principle exploit) while a negative $\bar{\rho}$ means subgroups diverge, creating structurally opposed signal that no single persona prompt can simultaneously satisfy.
To connect these dataset-side quantities to LLM alignment, we compute Spearman rank correlations between each quantity's per-attribute ranking and the corresponding single-component LLM alignment ranking. Full construction details for the interaction features and coherence analysis are in \Cref{app:word_demographic_interaction}.

\subsection{Neuron Probing}
\label{sec:neruon_setup}
To understand how demographic prompting affects internal model behavior, we probe neuron activations following the interpretability framework of \citet{ying-etal-2025-disentangling}. Their work traces neuron activations across \emph{languages} to explain cultural-linguistic synergy in multilingual LLMs. We adapt this to compare across \emph{prompting conditions} instead: for the same text in the same language, we compare which neurons fire when the model receives a demographic persona versus no persona (baseline).
Following \citet{geva2021transformer}, we define the $i$-th neuron at layer $l$ as the $i$-th element of the post-activation vector $\sigma(\mathbf{W}_{\text{up}}^l \cdot \mathbf{h}^l) \in \mathbb{R}^{d_m}$, where $\sigma$ is the MLP activation function (SiLU, GELU, etc.). We register forward hooks on the activation function of every MLP block to record these values. We only consider activations at \emph{response token positions} (i.e., from where the model's generated answer begins to where it ends) since our interest is in how demographic context changes generation behavior, not prompt encoding.

\paragraph{Key Neuron selection.}
For each sample, we select the top-$k$ neurons per layer by activation magnitude across all response positions, giving us a Key Neuron set $N_{\text{sample}}$:
\begin{equation}
N_{\text{sample}} = \bigl\{(i, l) \;\big|\;
  v^{r_t}_{(i,l)} \geq V_l^{\text{top-}k},\;
  r_t \in R\bigr\}
\label{eq:key_neuron}
\end{equation}
where $V_l^{\text{top-}k}$ is the $k$-th largest activation at layer $l$, $R = \{r_1, \ldots, r_n\}$ are the response tokens, and $v^{r_t}_{(i,l)}$ is the activation of neuron $i$ at layer $l$ for token $r_t$. Following the ablation study (in \Cref{app:ablation}), we evaluated threshold values of $k \in \{5, 10, 15, 20, 30, 40, 50, 100\}$ and found that selecting the top-10 neurons ($k$ = 10) most effectively captures the model's knowledge for our given task and thus set $k$ = 10 for our main results.
%We sweep $k \in \{5, 10, 15, 20\}$ and report $k = 10$ in the main results as we found that.\mknote{Need justification why we use top-k=10}

% \paragraph{Specialized neuron set.}
% Each base text appears under both a baseline (no-demographic) prompt and one or more identity-conditioned prompts. For a matched pair sharing the same base text, the \emph{specialized neuron set} is:
% %
% \begin{equation}
% S = N_{\text{demo}} \setminus N_{\text{baseline}}
% \label{eq:specialized}
% \end{equation}
% %
% i.e., the neurons active under the demographic prompt but not under the baseline. The \emph{specialization proportion} is:
% %
% \begin{equation}
% p = \frac{|N_{\text{demo}} \setminus N_{\text{baseline}}|}{|N_{\text{demo}}|}
% \label{eq:spec_prop}
% \end{equation}
% %
% A high $p$ means the demographic prompt engages a substantially different set of neurons. We compute $p$ at the individual pair level and then average across all matched pairs within each (task, model, attribute) combination. 

\paragraph{Specialized neuron set.}
Each base text appears under both a baseline (no-demographic) prompt and one or more identity-conditioned prompts. For a matched pair sharing the same base text, the \emph{specialized neuron set} is $S = N_{\text{demo}} \setminus N_{\text{baseline}}$, i.e., the neurons active under the demographic prompt but not under the baseline. The \emph{specialization proportion} is:
\begin{equation}
p = \frac{|N_{\text{demo}} \setminus N_{\text{baseline}}|}{|N_{\text{demo}}|}
\label{eq:spec_prop}
\end{equation}
A high $p$ means the demographic prompt engages a substantially different set of neurons. We compute $p$ at the individual pair level and then average across all matched pairs within each (task, model, attribute) combination.

\section{Results and Discussion}

\subsection{Combinatorial Effects of Demographic Attributes on LLM Alignment (RQ1)}
\label{sec:rq1_dis}

\begin{figure}[t]
\centering
\includegraphics[width=1.0\linewidth]{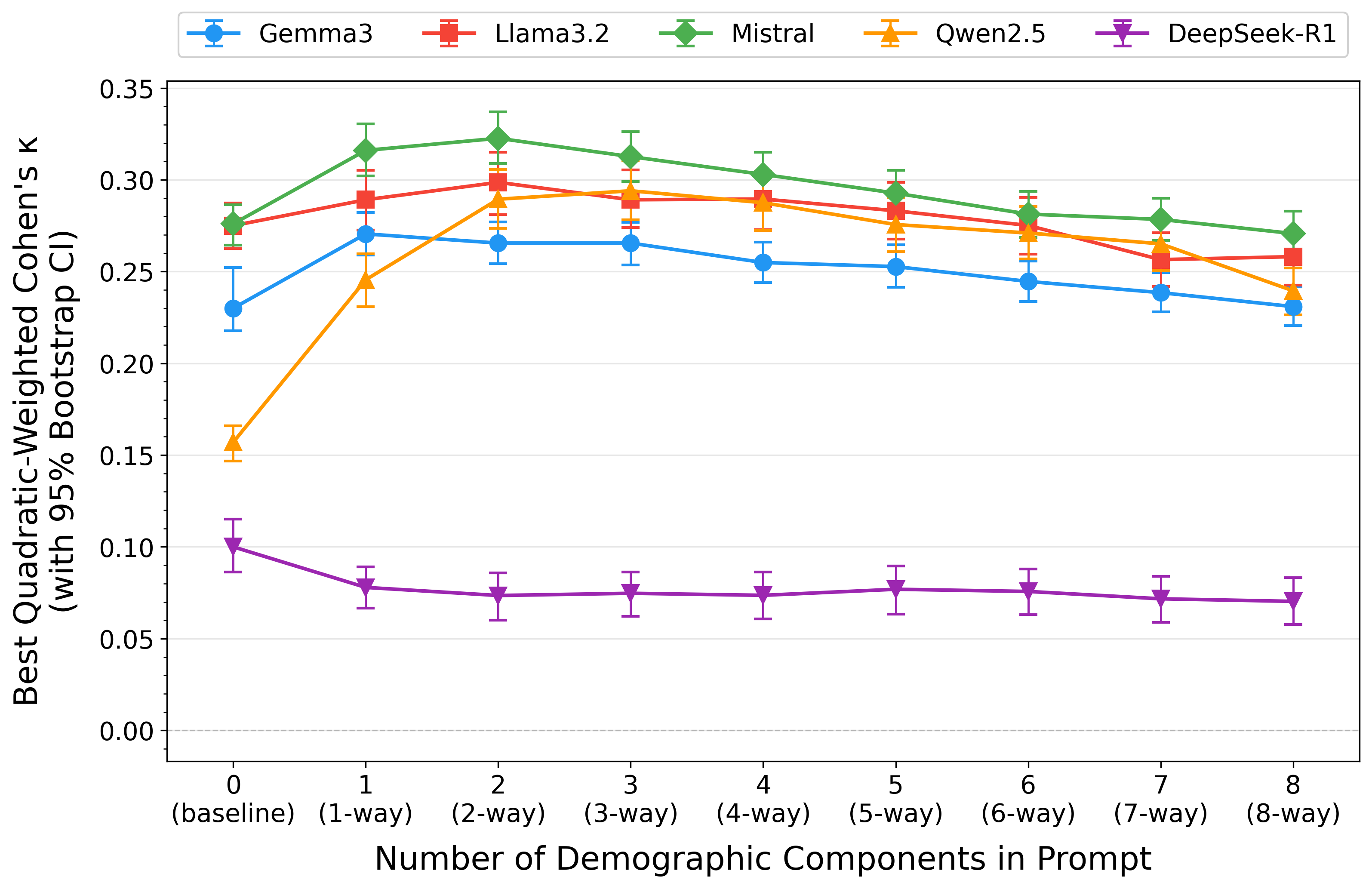}
\caption{Best quadratic-weighted Cohen's~$\kappa$ on the \textbf{Toxicity task} by the number of demographic prompt components across five LLMs.}
\label{fig:example_toxicity}
\end{figure}

\begin{figure}[t]
\centering
\includegraphics[width=1.0\linewidth]{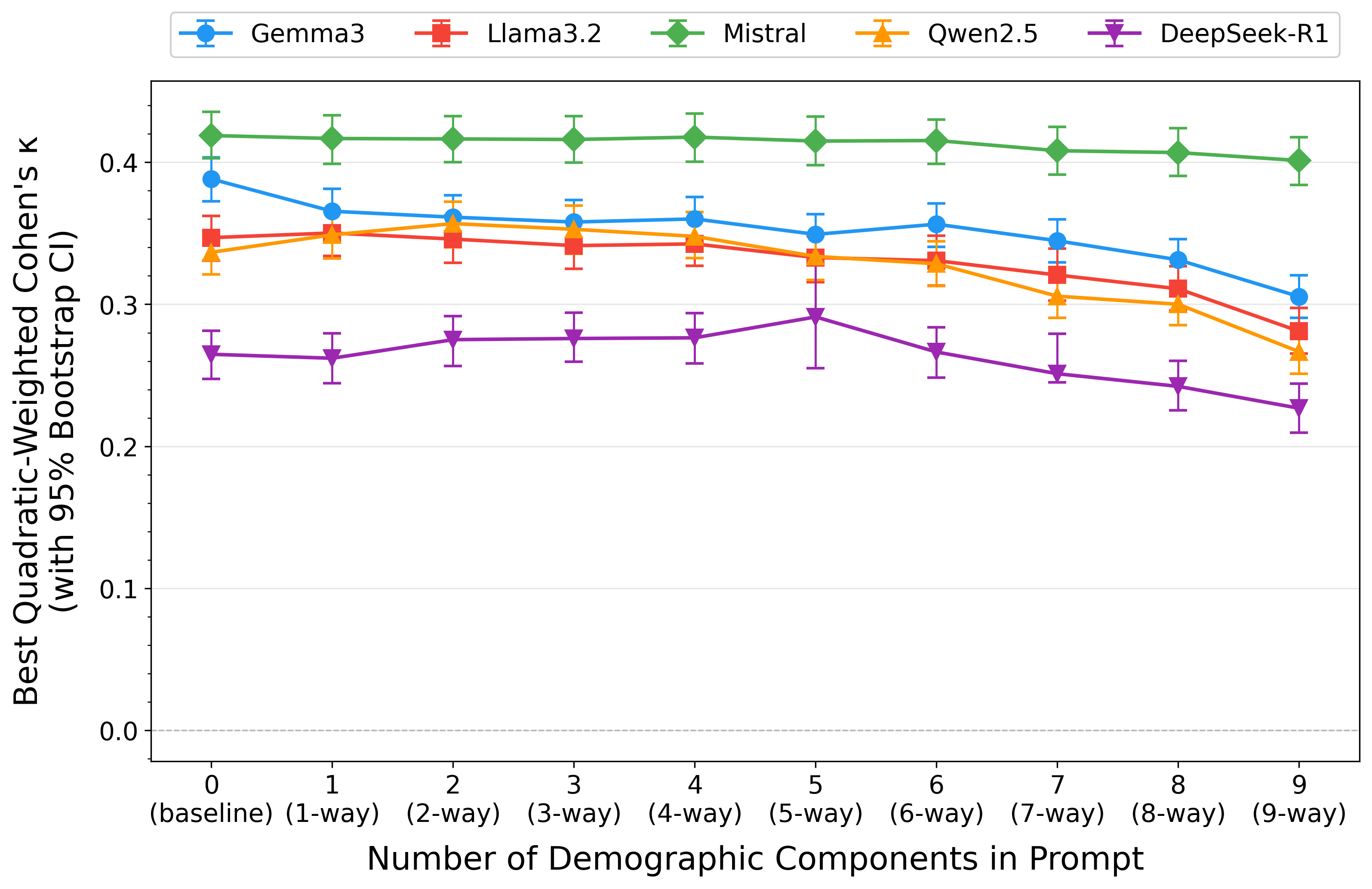}
\caption{Best quadratic-weighted Cohen's~$\kappa$ on the \textbf{Sentiment task} by the number of demographic prompt components across five LLMs.}
\label{fig:example_sentiment}
\end{figure}

\begin{figure}[t]
\centering
\includegraphics[width=1.0\linewidth]{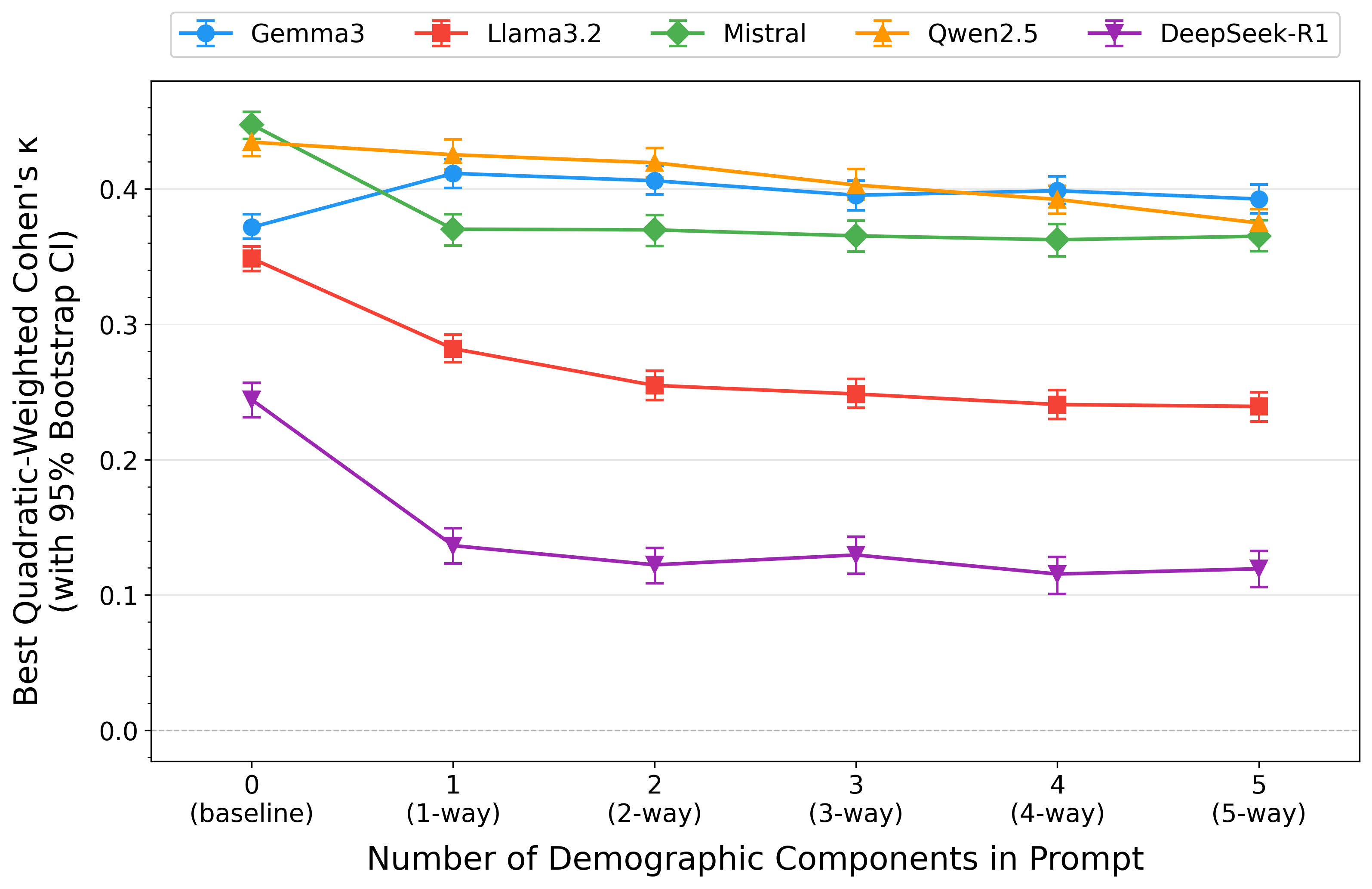}
\caption{Best quadratic-weighted Cohen's~$\kappa$ on the \textbf{Politeness task} by the number of demographic prompt components across five LLMs.}
\label{fig:example_politeness}
\end{figure}

\begin{figure}[t]
\centering
\includegraphics[width=1.0\linewidth]{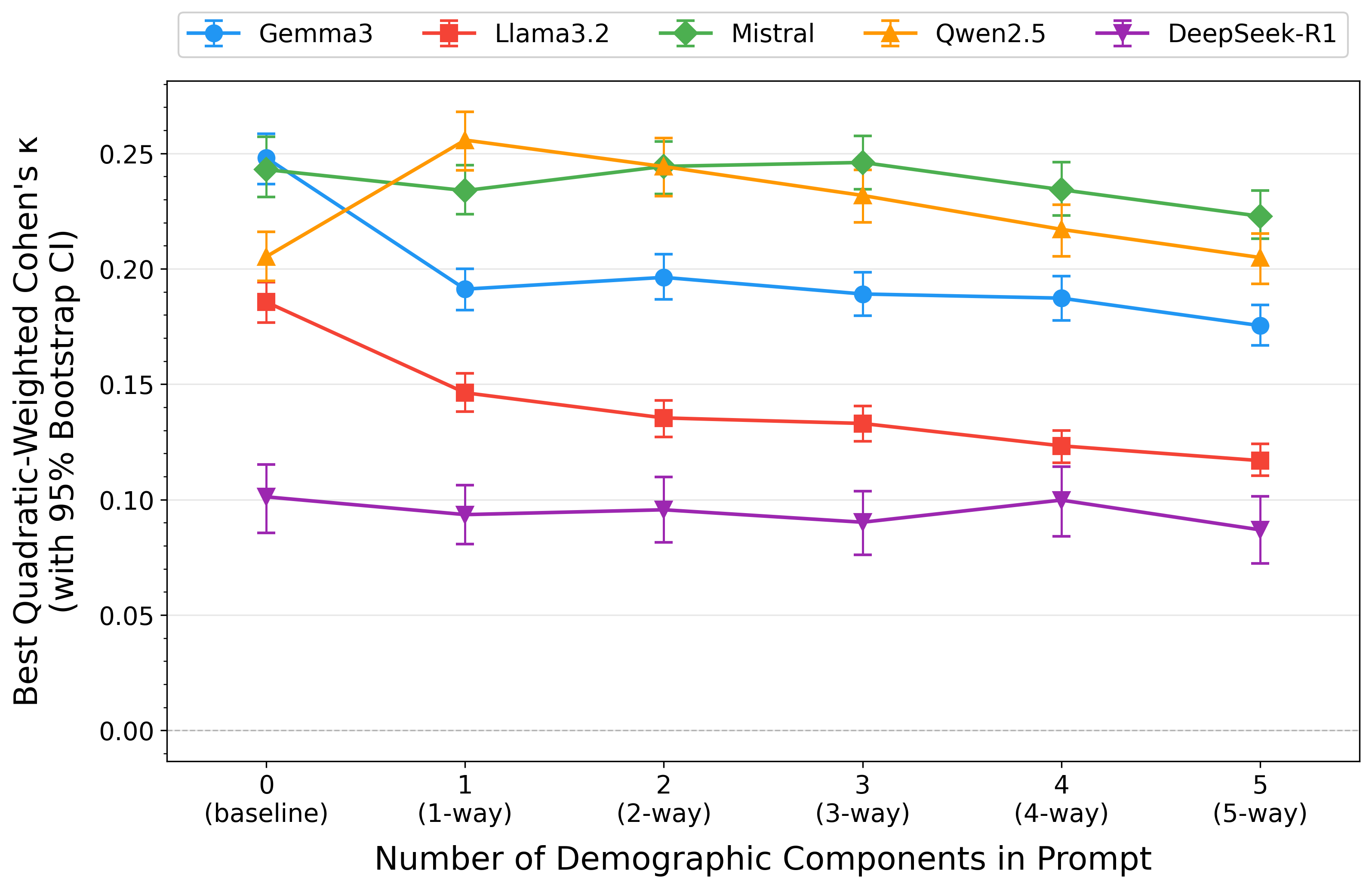}
\caption{Best quadratic-weighted Cohen's~$\kappa$ on the \textbf{Offensiveness task} by the number of demographic prompt components across five LLMs.}
\label{fig:example_offensiveness}
\end{figure}

\begin{figure}[t]
\centering
\includegraphics[width=1.0\linewidth]{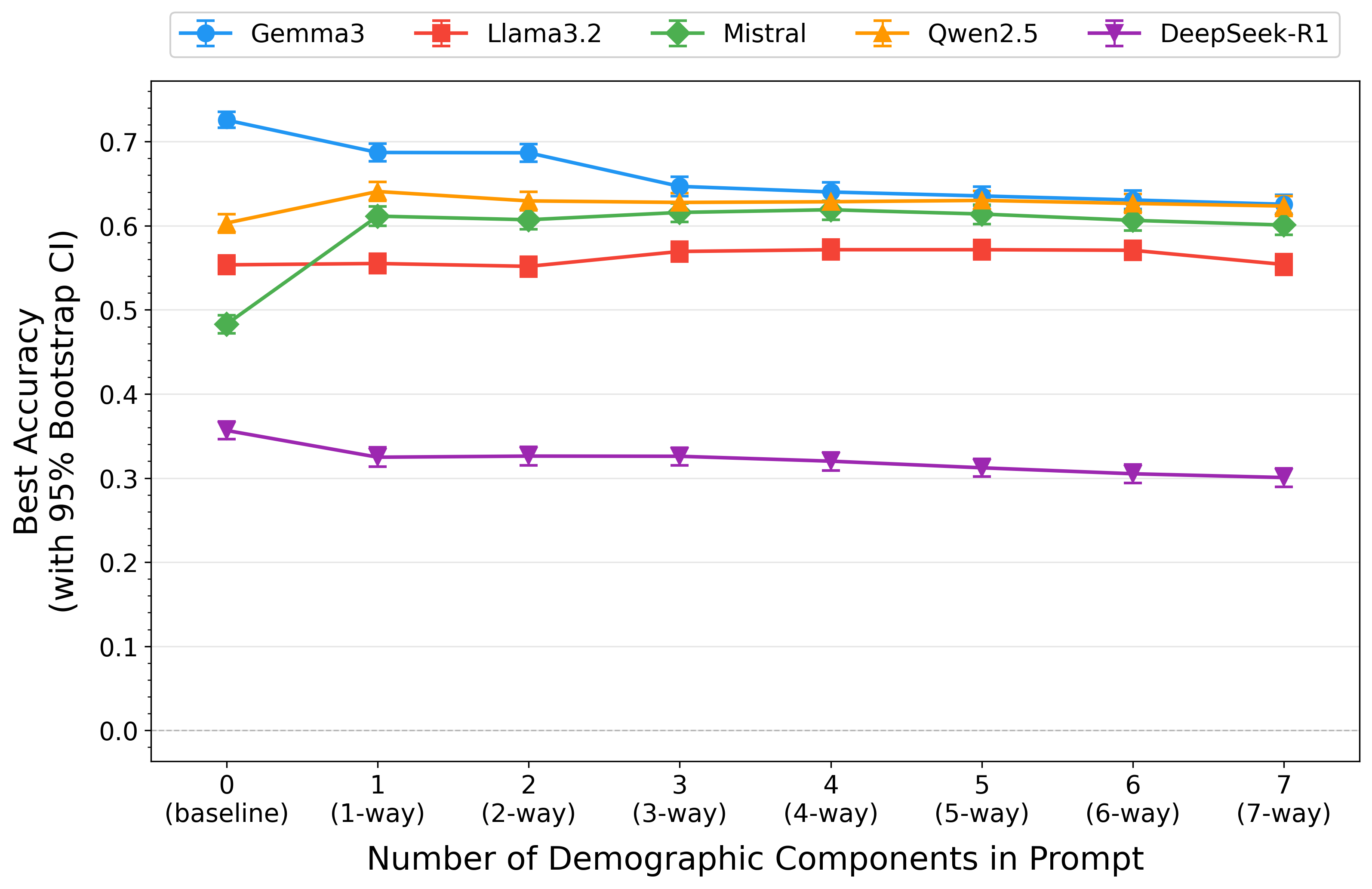}
\caption{Best accuracy on the \textbf{Emotion task} by the number of demographic prompt components across five LLMs.}
\label{fig:example_emotion}
\end{figure}

Across all five tasks, a consistent structural finding emerges: \textbf{\textit{alignment peaks with a small number of high-signal demographic components and degrades when prompted with the full attribute set.}} The optimal number of components, which specific attributes matter, and whether demographic prompting helps or hurts at all, however, \textbf{vary substantially across models and tasks.} Figures~\ref{fig:example_toxicity}–\ref{fig:example_emotion} trace the best achievable Cohen's~$\kappa$ (or accuracy for Emotion) with the number of demographic components in the prompt; full per-model breakdowns are in \Cref{tab:combined_toxicity_kappa_components,tab:combined_sentiment_kappa_components,tab:combined_politeness_kappa_components,tab:combined_offensiveness_kappa_components,tab:combined_emotion_kappa_components} (\Cref{app:detail_model_results}).

\paragraph{Toxicity.}
Toxicity is the task on which demographic prompting is most broadly beneficial. As \Cref{fig:example_toxicity} shows, four of the five models peak above their no-demographic baseline with a compact set of attributes: Gemma with \texttt{religion\_important} alone ($\kappa = 0.271$); Llama with \texttt{lgbtq\_status} + \texttt{race} ($\kappa = 0.299$); Qwen with \texttt{age\_range} + \texttt{lgbtq\_status} + \texttt{religion\_important} ($\kappa = 0.294$); and Mistral with \texttt{lgbtq\_status} + \texttt{political\_affiliation} ($\kappa = 0.323$, a gain of nearly five points over its baseline of $0.276$). DeepSeek is the sole exception: its baseline represents its highest performance (see \Cref{tab:combined_toxicity_kappa_components}). Beyond the optimal point, extending the prompt consistently hurts: the 8-component \textit{``all together''} prompt is never the best configuration and, for most models, represents a statistically significant degradation (see \Cref{tab:combined_toxicity_kappa_components}).

\paragraph{Sentiment Analysis.}
Sentiment presents a more mixed picture (\Cref{fig:example_sentiment}), with demographic prompting offering only modest and model-specific benefits. The three models that do gain are Llama with \texttt{gender} alone, Qwen with \texttt{education} + \texttt{hispanic\_latino}, and DeepSeek with a complex 5-way combination. For Gemma and Mistral, no demographic prompt surpasses their respective baselines. The curves in \Cref{fig:example_sentiment} are largely flat or declining from the single-component level onward, \textbf{indicating a weak and model-dependent demographic signal.}

\paragraph{Politeness and Offensiveness.}
\textbf{Both tasks show broad resistance to demographic prompting.} On Politeness, four of the five models (all except Gemma) achieve their best results at baseline, with degradations that are not marginal: adding \texttt{race} drops Llama's $\kappa$ from $0.349$ to $0.252$ and DeepSeek's from $0.244$ to $0.101$ (see \Cref{fig:example_politeness}; \Cref{tab:combined_politeness_kappa_components}). Gemma is the sole exception, with every single-attribute prompt exceeding its baseline, though combining attributes degrades these gains. On \textbf{Offensiveness, \texttt{race} is the only attribute that can selectively improve alignment}: Qwen peaks with \texttt{race} alone, and \texttt{race} appears in every top Mistral configuration (see \Cref{fig:example_offensiveness}; \Cref{tab:combined_offensiveness_kappa_components}). For the remaining models, no combination surpasses the baseline, and introducing attributes such as \texttt{age} or \texttt{gender} produces consistent, often significant degradations.

\paragraph{Emotion Attribution.}
Emotion Attribution produces the sharpest polarization across models (see \Cref{fig:example_emotion}; \Cref{tab:combined_emotion_kappa_components}). \textbf{\texttt{age} dominates for the models that benefit}: Mistral improves by 12.9~pp and Qwen by 3.8~pp, while Llama gains only through specific 4- and 5-way combinations anchored on \texttt{age}, \texttt{gender}, and parental occupations (up to $0.572$ vs.\ baseline $0.560$). Conversely, Gemma and DeepSeek are degraded by every demographic prompt without exception. That \texttt{age} cannot help these two models, whose strong baselines appear robust to persona prompting, highlights that \textbf{the utility of any demographic attribute is conditioned on model architecture as much as on dataset signal.}

\subsection{The Role of Directional Coherence in Demographic Prompting (RQ2)}
\label{sec:directional_coherence}

%\begin{center}
%\fbox{%
%    \parbox{0.455\textwidth}{\textbf{\underline{RQ2}: \textit{Do datasets contain patterns that can be exploited or suggest demographic-level signal, and do LLM outputs align with human label in a way that reflects that demographic signal?}}}}%\vspace{2.5mm}
%\label{sec:rq2}
%\end{center}
%OR

% \begin{center}
% \fbox{%
%     \parbox{0.455\textwidth}{\textbf{\underline{RQ2}: \textit{To what extent do dataset-level demographic signals predict when demographic prompting improves LLM–human alignment, and do LLM outputs reflect the learnability and directional coherence of those signals?}}}}%\vspace{2.5mm}
% \label{sec:rq2}

%\end{center}
% \gknote{Change the wording for this last part ``how do these signals affect the LLM behavior and response'' something along the lines of ``and do we find evidence that the LLM outputs reflect that structure?''}

\begin{table}[htbp]
\centering
\small
\resizebox{\columnwidth}{!}{%
\setlength{\tabcolsep}{4.0pt}
\begin{tabular}{lrrrrr}
\toprule
\textbf{Model} & \textbf{Toxicity} & \textbf{Sentiment} & \textbf{Politeness} & \textbf{Offensiveness} & \textbf{Emotion} \\
\midrule
Gemma             &  0.261 & -0.234 & -0.300 & -0.100 &  0.178 \\
Llama             & -0.333 & -0.301 & -0.400 &  0.400 & -0.214 \\
Qwen              &  0.023 & -0.283 & -0.100 & -0.500 &  0.250 \\
DeepSeek          &  0.571 & -0.108 & -0.400 &  0.100 &  0.285 \\
Mistral           & -0.595 &  0.267 & -0.200 & -0.100 & -0.285 \\
% \midrule
% Mean (all models) & $-0.0952$ & $-0.1833$ & $-0.3000$ & $-0.1000$ &  0.1786 \\
\bottomrule
\end{tabular}%
}
\caption{Spearman rank correlation results ($\rho$) for \textbf{SHAP importance vs. LLM performance}. We compare Cohen's kappa ($\kappa$) for all tasks except Emotion, for which we use accuracy. None of the settings are statistically significant.}
\label{tab:spearamn_shap_VS_models_alignment}
\end{table}

%%%%%% Pearson correlation results (r) for LSVC performance vs. LLM performance across different datasets%%%%%

% \begin{table*}[htbp]
% \centering
% \small
% \begin{tabular}{lrrrrr}
% \toprule
% \textbf{Model} & \textbf{Toxicity} & \textbf{Sentiment} & \textbf{Politeness} & \textbf{Offensiveness} & \textbf{Emotion} \\
% \midrule
% Gemma    & 0.2603  & 0.6483  & 0.6448  & $-0.6682$ & $-0.5697$ \\
% Llama    & 0.0467  & 0.6488  & 0.3336  & $-0.6982$ & $-0.7199$ \\
% Qwen     & 0.5382  & 0.2338  & 0.6867  & $-0.6801$ & 0.2989 \\
% DeepSeek & $-0.3163$ & 0.3933  & $-0.2057$ & 0.8720  & 0.4805 \\
% Mistral  & $-0.0064$ & 0.3854  & 0.1710  & $-0.8262$ & $-0.4944$ \\
% \bottomrule
% \end{tabular}
% \caption{Pearson correlation results ($r$) for LSVC performance vs. LLM performance across different datasets. No values were statistically significant ($p < 0.05$).}
% \label{tab:lsvc_pearson_results}
% \end{table*}

Using the setup described in \Cref{sec:dataset_singal_Setup}, we organize the RQ2 findings around two contrasts: whether the \emph{magnitude} of demographic influence (SHAP, \Cref{tab:shap_all_tasks}) or the \emph{learnability} of demographic signal (LSVC $\kappa$, \Cref{tab:lsvc_performance_all}) predicts LLM alignment, and connect both to the \emph{directional coherence} of each attribute's annotation signal (Fisher-averaged $\bar{\rho}$, in \Cref{tab:fisher_rhobar} (full results in Tables~\ref{tab:spearman_summary_toxicity}--\ref{tab:spearman_summary_emotion})).
% \footnote{We assess whether LLM outputs reflect dataset-side demographic structure indirectly, by testing whether attributes that are more influential, more learnable, or more directionally coherent in the human annotation data are also the attributes that yield better alignment when used as prompt cues.}

\paragraph{SHAP importance does not predict LLM alignment.}
Table~\ref{tab:spearamn_shap_VS_models_alignment} reports Spearman rank correlations between the normalized Mean~$|\text{SHAP}|$ importance of each single-attribute demographic feature and the corresponding single-component Cohen's~$\kappa$ (or accuracy for Emotion) achieved by each model.
None of the 25 (model~$\times$~task) correlations reaches statistical significance ($p < 0.05$).
The $\rho$ values scatter widely from -0.595 (Mistral, Toxicity) to +0.571 (DeepSeek, Toxicity) with no consistent pattern.
The implication is that \textbf{knowing which demographic attributes most strongly predict variation in human annotations provides no reliable information about which attributes will improve a given LLM's alignment when used as a prompt cue}.

\begin{table}[htbp]
\centering
\small
\resizebox{\columnwidth}{!}{%
\setlength{\tabcolsep}{4.0pt}
\begin{tabular}{lrrrrr}
\toprule
\textbf{Model} & \textbf{Toxicity} & \textbf{Sentiment} & \textbf{Politeness} & \textbf{Offensiveness} & \textbf{Emotion} \\
\midrule
Gemma    & 0.523  & 0.426  & 0.400  & -0.700 & -0.522 \\
Llama    & 0.071  & 0.376  & 0.200  & -0.700 & -0.666 \\
Qwen     & \textbf{0.714} & 0.283  & 0.700  & -0.700 & 0.216 \\
DeepSeek & -0.190 & 0.502  & -0.500 & 0.700  & 0.342 \\
Mistral  & -0.047 & 0.175  & -0.100 & -0.800 & -0.540 \\
\bottomrule
\end{tabular}
}
\caption{Spearman rank correlation results ($\rho$) for \textbf{LSVC performance vs. LLM performance}. We compare Cohen's kappa ($\kappa$) for all tasks except Emotion, for which we use accuracy. Statistically significant values ($p < 0.05$) are indicated in \textbf{bold}.}
\label{tab:lsvc_spearman_results}
\end{table}

\paragraph{LSVC learnability appears to be a more informative (though task-dependent) predictor.}
Table~\ref{tab:lsvc_spearman_results} reports the Spearman rank correlation between LSVC Cohen's~$\kappa$ (or accuracy for Emotion) and the corresponding single-component LLM $\kappa$ (or accuracy). 
% These LSVC scores capture how well a classifier can learn to predict the task label using word--demographic interaction features for each attribute; the Spearman $\rho$ tells us whether more learnable attributes also tend to produce better LLM alignment.
On Toxicity, the pattern is predominantly positive: Gemma ($\rho = 0.523$), \textbf{Qwen ($\rho = 0.714$, $p < 0.05$)}, and Mistral (-0.047) near zero. Concretely, the three attributes with the highest LSVC $\kappa$ on Toxicity (Table~\ref{tab:lsvc_performance_all}) \texttt{education} (0.201), \texttt{lgbtq\_status} (0.200), and \texttt{race} (0.186) are the same attributes yielding Qwen's strongest single-component alignment gains ($\kappa = 0.198$, 0.246, and 0.172, vs.\ baseline 0.157).
On Sentiment, all five correlations are positive (0.176--0.502), suggesting a weak-to-moderate tendency for more learnable attributes to benefit alignment, though none reaches significance.
Two systematic negative patterns emerge. On Offensiveness, four of the five models produce $\rho \leq -0.700$ (Gemma, Llama, Qwen: -0.700; Mistral: -0.800), meaning that the most learnable attributes tend to yield the worst LLM alignment. This inversion is concrete: \texttt{age}, the highest-LSVC attribute (0.316, Table~\ref{tab:lsvc_performance_all}), produces some of the largest alignment degradations (e.g., Gemma drops from 0.248 to 0.152; Llama from 0.186 to 0.131, Table~\ref{tab:combined_offensiveness_kappa_components}), while \texttt{race} ranked second-lowest in LSVC $\kappa$ (0.302) is the only attribute that significantly improves any model's alignment (Qwen: 0.256 vs.\ baseline 0.205).

\begin{table}[t]
\centering
\resizebox{\columnwidth}{!}{%
\begin{tabular}{lrrrrr}
\toprule
\textbf{Attribute} & \textbf{Toxicity} & \textbf{Sentiment} & \textbf{Politeness} & \textbf{Offensiveness} & \textbf{Emotion} \\
\midrule
age            & +0.125 & +0.032 & +0.048 & +0.139 & +0.280 \\
race           & +0.250 & +0.207 & +0.144 & +0.058 & ---     \\
gender         & -0.087 & -0.633 & -0.046 & -0.182 & -0.087 \\
lgbtq\_status  & +0.208 & ---     & ---     & ---     & ---     \\
\bottomrule
\end{tabular}%
}
\caption{Fisher-averaged $\bar{\rho}$ for selected demographic attributes across tasks. All values are significant. \textbf{This table reports only the attributes referenced in the main discussion; complete per-attribute results are in Tables~\ref{tab:spearman_summary_toxicity}--\ref{tab:spearman_summary_emotion} (Appendix~\ref{app:rq2_detail}).}}
\label{tab:fisher_rhobar}
\end{table}

\paragraph{Directional coherence explains when learnability helps versus hurts.}
The Fisher-averaged $\bar{\rho}$ values (\Cref{tab:fisher_rhobar} and Tables~\ref{tab:spearman_summary_toxicity}--\ref{tab:spearman_summary_emotion})
% \gknote{The way this is written, I think reviewers will complain that the results rely heavily on the Appendix. Restructure so that the table references are supplementary, alongside the reference to the Appendix at the end of this paragraph.} 
resolve the apparent contradiction between positive and negative LSVC correlations across tasks.
On Toxicity, the attributes with the most coherent directional signals, \texttt{race} ($\bar{\rho} = +0.250$) and \texttt{lgbtq\_status} ($\bar{\rho} = +0.208$) (\Cref{tab:fisher_rhobar}) are precisely those driving alignment gains in Table~\ref{tab:combined_toxicity_kappa_components}, and their high LSVC $\kappa$ values contribute to the positive Spearman correlations in Table~\ref{tab:lsvc_spearman_results}.
On Offensiveness, however, the pattern inverts. Although \texttt{age} is the most learnable attribute (LSVC $\kappa = 0.316$), its moderate Fisher $\bar{\rho}$ (+0.139) (\Cref{tab:fisher_rhobar}) reflects lexical patterns a classifier can exploit but a simple persona prompt cannot, learnability here reflects classifier capacity, not prompt-exploitable structure. \texttt{Gender}'s negative $\bar{\rho}$ (-0.182; Man vs.\ Woman $\rho = -0.319$) means its subgroups disagree on which words signal offensiveness, making persona-based alignment structurally challenging. Together, these produce the strong negative Spearman correlations in Table~\ref{tab:lsvc_spearman_results}.
The same mechanism applies to Sentiment, where \texttt{gender} carries the most severely opposed signal in the entire study (Fisher $\bar{\rho} = -0.633$; Female vs.\ Male $\rho = -0.633$), explaining why gender-based prompts fail to produce alignment gains despite moderate SHAP importance (0.107, Table~\ref{tab:shap_all_tasks}) and LSVC learnability ($\kappa = 0.203$, Table~\ref{tab:lsvc_performance_all}). Detailed per-attribute breakdowns are in Appendix~\ref{app:rq2_detail}.

% \paragraph{Summary answer to RQ2.}
% The dataset contains exploitable demographic-level signal, but that signal's relationship to LLM alignment depends on its nature, not just its magnitude.
% At the first level, the overall strength of demographic influence on human annotations (SHAP) is uninformative: no significant SHAP--alignment correlation was found for any model on any task.
% At the second level, the learnability of word--demographic interaction patterns (LSVC $\kappa$) is a meaningfully better predictor, as shown by the significant positive correlation for Qwen on Toxicity ($\rho = 0.714$) and the broadly positive trends on Sentiment.
% However, learnability \emph{inverts} as a predictor on Offensiveness and Emotion, where the most learnable attributes carry directionally opposed subgroup signals (negative Fisher $\bar{\rho}$) that no single persona prompt can resolve.\gknote{The summary preceding this note should be dramatically shortened. I think the non-summary portion is easy enough to follow.}
% The overall conclusion is that LLM demographic alignment cannot be predicted from any single dataset-side quantity; it requires jointly characterizing both the learnability \emph{and} the directional coherence of each attribute's annotation signal.

\subsection{Neuron Probing (RQ3)}
\label{sec:rq3_dis}

\paragraph{Rationale and Interpretation Framework.}

Our neuron probing analysis is grounded in interpretability findings showing that neurons with higher activation values during inference indicate knowledge usage relevant to the input~\citep{hong-etal-2025-intrinsic,cao2025model,zhao2024large,tang-etal-2024-language}. Building on this, \citet{ying-etal-2025-disentangling} propose that models activating a larger proportion of specialized neurons for a given context demonstrate stronger knowledge utilization and better task performance. Following this reasoning, we test the \textbf{hypothesis} that \textit{ a higher proportion of specialized neurons triggered by a demographic cue reflects deeper internal engagement with that attribute, potentially steering the model toward the corresponding annotation norms and improving alignment scores ($\kappa$ or accuracy)}. We treat this as a testable hypothesis rather than an assumption. To our knowledge, neuron probing has not previously been applied to demographic alignment in subjective annotation tasks. Using the setup described in \Cref{sec:neruon_setup}, we compute specialized neurons for single-attribute prompts and report full results in Tables~\ref{tab:specialized_toxicity}--\ref{tab:specialized_emotion} (\Cref{app:neuron_probing}). We then measure the Pearson correlation (following \citet{ying-etal-2025-disentangling}) between each attribute’s mean specialization proportion and its alignment score ($\kappa$ or accuracy), reported in \Cref{tab:neuron_count_correlation}, to examine whether attributes activating more specialized neurons also achieve stronger alignment with human annotations.

\begin{table}[htbp]
\centering
\small
\resizebox{\columnwidth}{!}{%
\setlength{\tabcolsep}{4.0pt}
\begin{tabular}{lrrrrr}
\toprule
\textbf{Model} & \textbf{Toxicity} & \textbf{Sentiment} & \textbf{Politeness} & \textbf{Offensiveness} & \textbf{Emotion} \\
\midrule
Gemma & -0.222 & \textbf{-0.888} & -0.823 & -0.782 & \textbf{-0.982} \\
Llama & \textbf{0.741} & -0.657 & 0.052 & -0.120 & \textbf{-0.772} \\
Qwen & \textbf{0.769} & -0.608 & 0.118 & 0.432 & -0.318 \\
DeepSeek & 0.706 & -0.049 & -0.491 & 0.682 & 0.033 \\
Mistral & 0.660 & \textbf{-0.727} & 0.731 & 0.345 & 0.678 \\
\bottomrule
\end{tabular}
}
\caption{Pearson correlation between the number of neurons and LLM performance/alignment metrics. We compare Cohen's kappa ($\kappa$) for all tasks except Emotion, for which we use accuracy. Values in \textbf{bold} are statistically significant ($p < 0.05$).}
\label{tab:neuron_count_correlation}
\end{table}

\paragraph{Selective support and the high-volume paradox.}
The specialized-neuron hypothesis receives partial support on Toxicity. 
Table~\ref{tab:neuron_count_correlation} shows strong positive correlations for 
Llama ($r=0.742$) and Qwen ($r=0.769$), indicating that demographic attributes 
that activate more specialized neurons tend to produce better alignment with 
human toxicity annotations. Attributes such as \texttt{lgbtq\_status} and 
\texttt{religion\_important} illustrate this pattern for Qwen: they activate the highest 
proportions of specialized neurons (see \Cref{tab:specialized_toxicity}) and also yield the largest alignment gains (see \Cref{tab:combined_toxicity_kappa_components}). 
Importantly, these attributes also exhibit directionally coherent lexical 
signals (Section~\ref{sec:directional_coherence}), suggesting that specialized 
neuron activation correlates with alignment only when the underlying annotation 
signal is structurally exploitable. However, this relationship does not hold 
universally. DeepSeek activates the largest number of specialized neurons 
across tasks (see \Cref{tab:specialized_toxicity,tab:specialized_sentiment,tab:specialized_politeness,tab:specialized_offensiveness,tab:specialized_emotion}), yet its no-demographic baseline consistently performs best and 
no demographic attribute improves performance. \textbf{This high-volume paradox 
suggests that neuron quantity alone does not reflect useful knowledge 
engagement; instead, widespread activation may indicate broad representational 
perturbation rather than targeted alignment-relevant processing.}

% \textbf{Task-dependent breakdown and signal coherence.}
% In several model–task combinations, the neuron–performance relationship 
% reverses. Significant negative correlations appear on Sentiment (Gemma 
% $r=-0.888$, Mistral $r=-0.727$) and Emotion (Gemma $r=-0.983$, Llama 
% $r=-0.772$). Here, attributes that activate more specialized neurons often 
% degrade performance, suggesting interference with already well-calibrated 
% representations. For example, in Sentiment, Gemma’s most activating attributes 
% (\texttt{political\_id}, \texttt{hispanic\_latino}) produce the largest 
% alignment drops, while \texttt{gender} activates the fewest neurons and 
% causes the smallest degradation despite carrying the most conflicting 
% annotation signal. A similar pattern appears in Emotion, where attributes 
% triggering the most new neurons lead to the steepest performance declines. 
% Even in Gemma’s Politeness case, where several demographic prompts improve 
% over baseline, the best-performing attribute (\texttt{occupation}) activates 
% the fewest specialized neurons. \textbf{Across tasks, these results indicate 
% that specialized-neuron counts alone are an unreliable proxy for alignment 
% utility; the directional coherence of the underlying annotation signal is the 
% critical factor determining whether demographic prompting helps or harms 
% performance.} For more detailed discussion on this see \Cref{app:neuron_probing}.

\paragraph{Task-dependent breakdown and signal coherence.}
In several model--task combinations, the neuron--performance relationship reverses. Significant negative correlations appear on Sentiment (Gemma r=-0.888, Mistral r=-0.727) and Emotion (Gemma r=-0.983, Llama r=-0.772), where attributes activating more specialized neurons often degrade performance, suggesting interference with already well-calibrated representations. For instance, on Sentiment, Gemma's most activating attributes (\texttt{political\_id}, \texttt{hispanic\_latino}) produce the largest alignment drops, while \texttt{gender} activates the fewest neurons yet causes the smallest degradation despite carrying the most conflicting annotation signal. Similarly, on Politeness, Gemma's best-performing attribute (\texttt{occupation}) activates the fewest specialized neurons. \textbf{Across tasks, specialized-neuron counts alone are an unreliable proxy for alignment utility; the directional coherence of the underlying annotation signal is the critical factor determining whether demographic prompting helps or harms performance.} See \Cref{app:neuron_probing} for detailed discussion.

\section{Conclusion}

We presented a systematic study of demographic prompting across five tasks, enumerating all possible attribute combinations. Three findings stand out. First, alignment follows a consistent pattern: one to three high-signal attributes yield the best results, and the full attribute set never helps. Second, predicting which attributes will improve alignment requires jointly considering both learnability and directional coherence of each attribute's annotation signal. Third, specialized neuron activation correlates with alignment only when the underlying signal is structurally coherent, and activation volume alone does not index steerability. Collectively, these results demonstrate that demographic prompting is not a monolithic intervention, and that principled attribute selection grounded in signal quality is essential for effective use. Based on these findings, we provide task- and model-specific recommendations for practitioners in Appendix~\ref{sec:practical-suggestions}.

\section{Limitations}

\paragraph{English-only scope.}
All five datasets and our experimental setups are in English, so whether our findings on directional coherence and over-specification transfer to other languages or cultural settings remains an open question.

\paragraph{LLMs.}
All five LLMs range from 3B to 12B parameters. Larger models may exhibit different sensitivity to demographic prompts, and our findings may not generalize to proprietary models such as GPT-4 or Claude, which differ in training data, alignment procedures, and scale. We limited our tests to five models due to resource constraints and balancing the research budget.

\paragraph{Demographic category granularity.}
We treat demographic attributes as fixed categorical variables inherited from each dataset. This does not capture intersectional identities, within-category heterogeneity, or the fluid nature of certain demographic dimensions. Additionally, annotator pools in several datasets skew toward particular groups (e.g., predominantly White participants), which may limit the representativeness of subgroup-level analyses.

\paragraph{Prompt template sensitivity.}
Although we average results across three paraphrases of each prompting template to improve robustness, our findings remain conditioned on the general structure of persona-style prompts. Alternative prompting strategies (e.g., chain-of-thought or role-play framing) could yield different alignment patterns.

\paragraph{Correlational neuron analysis.}
Our neuron probing analysis is correlational rather than causal. We identify associations between specialized neuron activation and alignment outcomes, but \textit{we do not intervene on specific neurons to confirm a causal mechanism}.

\paragraph{Inter-annotator agreement in source datasets.}
We rely on original annotation labels as ground truth but do not report inter-annotator agreement (IAA) within the source datasets. Since subjective tasks are known to exhibit substantial annotator disagreement, low IAA would place an inherent ceiling on achievable alignment. Our $\kappa$ values should therefore be interpreted relative to this ceiling rather than as absolute measures of model quality.

\paragraph{Small-$n$ rank correlations and correlational scope.}
The Spearman correlations central to RQ2 (Tables~\ref{tab:spearamn_shap_VS_models_alignment} and~\ref{tab:lsvc_spearman_results}) are computed over as few as five matched attributes for Politeness and Offensiveness. At this sample size, individual rank swaps can substantially shift $\rho$, so the per-task correlations should be interpreted as indicative trends rather than precise effect-size estimates, though the consistent directionality across tasks and models lends support to our conclusions. Additionally, our RQ2 analysis is entirely correlational: we identify associations between dataset-side signal properties and LLM alignment but \textit{do not} manipulate these properties directly. Unmeasured confounds such as label distribution skew or text-level difficulty may partially account for the observed patterns, and we frame directional coherence as a diagnostic indicator rather than a confirmed causal mechanism.

\paragraph{Distillation effects on persona sensitivity} DeepSeek-R1-Distill-Qwen-7B was fine-tuned on chain-of-thought reasoning traces from the larger DeepSeek-R1 model. This distillation process may compress or discard the representational capacity needed to respond differentially to persona cues. Our high-volume paradox finding may therefore reflect a distillation artifact rather than a general property of demographic prompting, and we do not disentangle distillation effects from model scale or architecture in the current study.

\section*{Acknowledgements}

\bibliography{anthology,custom}

\appendix

\section{Dataset Details}
\label{app:detail_dataset}

\begin{table*}[ht]
\centering
\small
\begin{tabularx}{\textwidth}{l|l|X}
\hline
\textbf{Task} & \textbf{Dataset} & \textbf{Labels} \\ \hline
Toxicity & \citet{kumar2021designing} & Not at all toxic (52.02\%), Slightly toxic (19.15\%), Moderately toxic (12.95\%), Very toxic (9.48\%), Extremely toxic (6.40\%) \\ \hline
Sentiment & \citet{diaz2018addressing} & Neutral (42.88\%), Somewhat positive (23.47\%), Somewhat negative (19.22\%), Very positive (9.54\%), Very negative (4.89\%) \\ \hline
Politeness & \citet{pei-jurgens-2023-annotator} & somewhat polite (29.03\%), moderately polite (24.90\%), very polite (21.02\%), barely polite (13.88\%), not polite at all (11.17\%) \\ \hline
Offensiveness & \citet{pei-jurgens-2023-annotator} & not offensive at all (57.66\%), barely offensive (16.80\%), somewhat offensive (12.20\%), moderately offensive (7.97\%), very offensive (5.38\%) \\ \hline
Emotion & \citet{scherer1994evidence} & anger (14.29\%), sadness (14.29\%), disgust (14.29\%), shame (14.29\%), fear (14.29\%), joy (14.29\%), guilt (14.29\%) \\ \hline
\end{tabularx}
\caption{Label distribution for each task.}
\label{tab:my_task_labels}
\end{table*}

\begin{table*}[ht]
\centering
\small
\begin{tabularx}{\textwidth}{l|X}
\hline
\textbf{Attribute} & \textbf{Values (Percentage share)} \\ \hline
age\_range & 25 - 34 (39.51\%), 35 - 44 (25.00\%), 45 - 54 (13.11\%), 18 - 24 (11.77\%), 55 - 64 (7.33\%), 65 or older (3.10\%), Prefer not to say (0.16\%), Under 18 (0.02\%) \\\hline
education & Bachelor's degree (40.86\%), Some college (20.14\%), Master's degree (15.01\%), Associate degree (11.08\%), High school graduate (8.97\%), Professional degree (1.59\%), Doctoral degree (1.19\%), Less than high school (0.54\%), Prefer not to say (0.39\%), Other (0.18\%) \\\hline
gender & Female (51.64\%), Male (47.03\%), Unknown (0.78\%), Nonbinary (0.54\%) \\\hline
parental\_status & Yes (51.23\%), No (47.79\%), Prefer not to say (0.99\%) \\\hline
lgbtq\_status & Heterosexual (82.42\%), Bisexual (10.86\%), Homosexual (3.49\%), Prefer not to say (1.99\%), Other (0.86\%) \\\hline
political\_affiliation & Liberal (40.83\%), Conservative (26.69\%), Independent (26.30\%), Prefer not to say (4.04\%), Other (2.14\%) \\\hline
race & White (71.86\%), Black or African American (12.29\%), Asian (5.78\%), Hispanic (2.72\%), Prefer not to say (1.04\%), White/Hispanic (0.89\%), American Indian or Alaska Native (0.88\%), White/Black (0.81\%), Other (0.78\%), White/Asian (0.69\%), White/American Indian (0.59\%), White/Other (0.31\%), Missing (0.24\%), Native Hawaiian/Pacific Islander (0.18\%), Various mixed backgrounds (<0.12\% each) \\\hline
religion\_importance & Not important (31.79\%), Very important (31.44\%), Somewhat important (23.42\%), Not too important (12.00\%), Prefer not to say (1.36\%) \\ \hline
\end{tabularx}
\caption{Demographic characteristics and percentage shares for the \textbf{Toxicity} task dataset}
\label{tab:toxicity_attributes}
\end{table*}

Here, we provide a comprehensive overview of each of the datasets. \Cref{tab:my_task_labels} details the label distributions for each of the five tasks, capturing the class frequencies for Toxicity, Sentiment Analysis, Politeness, Offensiveness, and Emotion Attribution. We also provide specific demographic breakdowns including percentage shares for attributes such as age, race, gender, and education, across individual tables for Toxicity (\Cref{tab:toxicity_attributes}), Sentiment Analysis (\Cref{tab:sentiment_demographics}), Politeness (\Cref{tab:politeness_demographics}), Offensiveness (\Cref{tab:offensiveness_demographics}), and Emotion Attribution (\Cref{tab:emotion_demographics}). These summaries highlight the diversity of the rater pools whose judgments form the basis of our alignment evaluation.

1. \textbf{Toxicity:} The toxicity dataset was constructed from an initial corpus of 549,058 comments collected from Twitter, Reddit, and 4chan between December 2019 and August 2020. To address the natural class imbalance between benign and toxic content, \citet{kumar2021designing} employed a stratified sampling approach using scores from the Perspective API TOXICITY model. This method deliberately oversampled comments within score ranges that typically generate the highest rater disagreement. The resulting final dataset contains 107,620 comments, distributed across Twitter, 4chan, and Reddit. These comments were evaluated by a diverse group of 17,280 crowdsourced participants.

\begin{table*}[ht]
\centering
\small
\begin{tabularx}{\textwidth}{l|X}
\hline
\textbf{Attribute} & \textbf{Values (Percentage share)} \\ \hline
age & 60-69 (40.73\%), 50-59 (36.17\%), 70-79 (20.22\%), 80-89 (2.68\%), 90-99 (0.14\%), 100+ (0.06\%) \\ \hline
race & White (75.98\%), Black or African American (13.90\%), Asian (6.44\%), Other (2.10\%), American Indian or Alaska Native (1.22\%), Native Hawaiian or Pacific Islander (0.22\%), Middle Eastern (0.14\%) \\ \hline
hispanic\_latino & No (83.39\%), Yes (16.61\%) \\ \hline
income & \$50,000 - \$74,999 (20.80\%), \$35,000 - \$49,999 (15.83\%), \$25,000 - \$34,999 (13.77\%), \$75,000 - \$99,999 (13.68\%), \$15,000 - \$24,999 (10.87\%), \$100,000 - \$149,999 (9.26\%), Less than \$10,000 (5.54\%), \$10,000 - \$14,999 (5.18\%), More than \$200,000 (2.62\%), \$150,000 - \$199,999 (2.45\%) \\ \hline
education & Some college or associate's degree (37.94\%), Bachelor's degree (25.34\%), High school graduate/GED (19.70\%), Graduate or professional degree (15.08\%), Less than high school (1.94\%) \\ \hline
employment\_status & Retired (47.75\%), Working full-time (24.90\%), Working part-time (11.85\%), Unemployed (7.93\%), On disability (7.57\%) \\ \hline
living\_situation & Live with spouse/partner (51.98\%), Live alone (29.44\%), Live with family (16.84\%), Other (1.11\%), Nursing home (0.36\%), Assisted living (0.21\%), Retirement community (0.06\%) \\ \hline
political\_id & Moderate (38.80\%), Somewhat conservative (20.36\%), Somewhat liberal (17.09\%), Very conservative (14.90\%), Very liberal (8.85\%) \\ \hline
gender & Male (50.40\%), Female (49.54\%), Nonbinary (0.06\%) \\ \hline
\end{tabularx}
\caption{Demographic characteristics and percentage shares for the \textbf{Sentiment} task dataset.}
\label{tab:sentiment_demographics}
\end{table*}

\begin{table*}[ht]
\centering
\small
\begin{tabularx}{\textwidth}{l|X}
\hline
\textbf{Attribute} & \textbf{Values (Percentage share)} \\ \hline
race & White (72.55\%), Black or African American (12.65\%), Hispanic or Latino (6.88\%), Asian (6.32\%), Prefer not to disclose (0.79\%), Native Hawaiian or Pacific Islander (0.40\%), American India or Alaska Native (0.20\%), Hebrew (0.20\%) \\ \hline
age & $>$65 (13.48\%), 60-64 (11.06\%), 18-24 (10.94\%), 30-34 (10.76\%), 40-44 (10.29\%), 50-54 (9.29\%), 54-59 (9.22\%), 25-29 (8.71\%), 45-49 (8.31\%), 35-39 (7.55\%), Prefer not to disclose (0.40\%) \\ \hline
occupation & Employed (43.89\%), Retired (15.19\%), Self-employed (12.78\%), Unemployed (12.12\%), Homemaker (6.55\%), Student (5.37\%), Other (2.33\%), Prefer not to disclose (1.78\%) \\ \hline
education & College degree (46.95\%), High school diploma or equivalent (29.88\%), Graduate degree (19.00\%), Other (2.19\%), Prefer not to disclose (1.19\%), Less than a high school diploma (0.78\%) \\ \hline
gender & Woman (50.07\%), Man (46.74\%), Non-binary (2.60\%), Prefer not to disclose (0.60\%) \\ \hline
\end{tabularx}
\caption{Demographic attributes and percentage shares for the \textbf{Politeness} task dataset.}
\label{tab:politeness_demographics}
\end{table*}

2. \textbf{Sentiment:} The sentiment analysis dataset was constructed to evaluate age-related bias by isolating and manipulating age-identifying terms in naturalistic text. \citet{diaz2018addressing} initially collected a corpus by scraping 4,151 blog posts and 64,283 comments from a prominent ``elderblogger" community. From this text, they extracted sentences containing the word ``old" and applied strict exclusion criteria to isolate instances where the term specifically described people rather than objects. This filtering yielded a curated set of 121 base sentences. To establish a comparative baseline, the researchers duplicated these sentences and systematically replaced ``old" (and its variants) with ``young" (and its variants), resulting in a final standardized dataset of 242 paired sentences. Additionally, for broader synthetic testing and model retraining, the study generated a template-based dataset of around 135k sentences and isolated a subset of 13,781 age-related tweets filtered from the 1-million-tweet Sentiment140 corpus.

3. \textbf{Politeness and Offensiveness:} The POPQUORN dataset is derived from the Ruddit corpus \cite{hada-etal-2021-ruddit}, but it departs from the original Best–Worst Scaling (BWS) annotation method. Because BWS can struggle with skewed data and often produces label distributions that differ from standard rating scales \cite{louviere2015best}, the data was reannotated to better suit our experimental design. Specifically, instances were rescored on a 1-to-5 scale, ranging from 1 (not at all polite/offensive) to 5 (very polite/offensive). To ensure data quality and reduce noise, the MACE framework \cite{hovy-etal-2013-learning} was applied to filter out unreliable annotators. The refined dataset ultimately features responses from 1,484 individuals (complete with demographic metadata detailing their age, gender, race, education, and occupation), yielding 25,043 instances for the politeness task and 13,037 instances for the offensiveness task.

4. \textbf{Emotion:} ISEAR dataset includes 7,665 events of 7 emotion categories. They utilize information from 3000 respondents in the dataset covering 16 countries.

\begin{table*}[ht]
\centering
\small
\begin{tabularx}{\textwidth}{l|X}
\hline
\textbf{Attribute} & \textbf{Values (Percentage share)} \\ \hline
race & White (75.21\%), Black or African American (12.29\%), Asian (7.28\%), Native American (2.29\%), Hispanic or Latino (1.92\%), Arab American (0.38\%) \\ \hline
age & 54-59 (13.75\%), 35-39 (11.87\%), 18-24 (11.85\%), 30-34 (11.46\%), $>$65 (11.08\%), 25-29 (9.48\%), 40-44 (8.76\%), 45-49 (8.37\%), 50-54 (7.65\%), 60-64 (5.73\%) \\ \hline
occupation & Employed (48.52\%), Self-employed (12.98\%), Unemployed (12.92\%), Retired (11.09\%), Student (7.63\%), Homemaker (4.56\%), Other (1.92\%), Prefer not to disclose (0.38\%) \\ \hline
education & College degree (46.30\%), High school diploma or equivalent (30.56\%), Graduate degree (19.03\%), Other (2.29\%), Less than a high school diploma (1.82\%) \\ \hline
gender & Woman (50.02\%), Man (47.30\%), Non-binary (2.68\%) \\ \hline
\end{tabularx}
\caption{Demographic attributes and percentage shares for the \textbf{Offensiveness} task dataset.}
\label{tab:offensiveness_demographics}
\end{table*}

\begin{table*}[ht]
\centering
\small
\begin{tabularx}{\textwidth}{l|X}
\hline
\textbf{Attribute} & \textbf{Values (Percentage share)} \\ \hline
gender & female (54.78\%), male (45.09\%), Other (0.04\%) \\ \hline
father's\_occupation. & white collar academic (26.84\%), white collar nonacademic (19.31\%), self-employed nonacademic (15.13\%), blue collar untrained (11.32\%), blue collar trained (11.23\%), unemployed (7.39\%), self-employed academic (4.56\%), housewife (1.00\%), student (0.09\%) \\ \hline
mother's\_occupation. & housewife (43.96\%), white collar nonacademic (20.05\%), white collar academic (14.68\%), blue collar untrained (6.85\%), self-employed nonacademic (5.57\%), blue collar trained (4.29\%), unemployed (1.46\%), self-employed academic (0.64\%), student (0.27\%) \\ \hline
field\_of\_study & psychology (40.94\%), medical (16.98\%), social sciences (14.58\%), other (9.04\%), languages (5.93\%), law (4.47\%), natural science (4.38\%), engineering (1.46\%), fine arts (1.10\%) \\ \hline
country & zambia (10.04\%), sweden (7.39\%), china mainland (7.21\%), spain (7.12\%), bulgaria (6.66\%), malawi (6.57\%), finland (6.38\%), netherlands (6.30\%), austria (6.26\%), india (6.21\%), usa (5.55\%), new zealand (5.39\%), brazil (5.29\%), australia (5.29\%), honduras (5.01\%), norway (3.29\%) \\ \hline
age & 20.0 (16.69\%), 21.0 (15.24\%), 22.0 (12.92\%), 19.0 (12.14\%), 18.0 (8.66\%), 23.0 (7.85\%), 24.0 (5.84\%), 25.0 (3.82\%), 26.0 (3.19\%), 27.0 (2.74\%), 28.0 (2.46\%), 30.0 (2.01\%), 33.0 (1.28\%), 29.0 (1.19\%), 35.0 (1.19\%), 32.0 (1.10\%), 34.0 (0.82\%), 31.0 (0.82\%) \\ \hline
religion & catholic (32.74\%), protestant (26.80\%), areligious (24.28\%), hindu (5.66\%), others (3.65\%), native (3.01\%), jewish (0.53\%), buddhist (0.37\%) \\ \hline
\end{tabularx}
\caption{Demographic characteristics and percentage shares for the \textbf{Emotion} task dataset.}
\label{tab:emotion_demographics}
\end{table*}

\section{LLM}
\label{app:llm_detail}

For our analysis, we utilize five instruction-tuned language models sourced directly from the Hugging Face\footnote{\url{https://huggingface.co/}}: (1) \textbf{Llama-3.2-3B-Instruct} (\textit{meta-llama/Llama-3.2-3B-Instruct}); (2) \textbf{Mistral-7B-Instruct-v0.3} (\textit{mistralai/Mistral-7B-Instruct-v0.3}); (3) \textbf{Gemma-3-12B-it} (\textit{google/gemma-3-12b-it}); (4) \textbf{Qwen2.5-7B-Instruct} (\textit{Qwen/Qwen2.5-7B-Instruct}); and (5) \textbf{DeepSeek-R1-Distill-Qwen-7B} (\textit{deepseek-ai/DeepSeek-R1-Distill-Qwen-7B}).
We intentionally include a diverse selection of architectures, featuring both Western-trained models, such as Llama (Meta), Gemma (Google), and Mistral (Mistral AI), alongside Chinese-trained models, such as Qwen (Alibaba Cloud) and DeepSeek. This selection allows for a balanced assessment across different training methodologies and cultural data influences. To accommodate budget constraints and facilitate our available computational capabilities, we focused on these specific model scales. We used all the default hyperparameters for these models.

\section{Prompt Details}
\label{app:detail_prompt}

In Table \ref{tab:prompt-templates}, we present the specific prompting templates used for both the baseline evaluation and the with-demographic assessment, including two rephrased variants for each to ensure results are not artifact-dependent. In these templates, the placeholder \textbf{\texttt{{text}}} refers to the specific content the model is tasked to classify, while \textbf{\texttt{{task}}} defines the classification dimension, such as toxicity levels or sentiment polarity. The \textbf{\texttt{{options\_string}}} placeholder represents the set of labels corresponding to the specific task; for instance, a toxicity task include a 5-point scale ranging from ``Not at all toxic" to ``Extremely toxic." Crucially, the order of these options is randomized for every prompt. As prior studies \cite{zeng2025order,shi2025judging} have demonstrated that large language models are susceptible to ordering bias, often showing a preference for labels appearing earlier in a list, this randomization ensures that our results reflect the model's actual judgment rather than a structural artifact of the prompt. For example, a baseline prompt for a toxicity task might appear as: \textit{``Given the following text: 'He's lucky get didn't get a bullet put in his head', rate its overall toxicity. Respond using one of the following options: Not at all toxic, Slightly toxic, Moderately toxic, Very toxic, Extremely toxic. Respond with an answer in English,
without any explanation"} By shuffling these options in subsequent iterations, we mitigate positional preference and increase the robustness of our findings.

\begin{table*}[h!]
\centering
\small
\begin{tabular}{@{}llp{10cm}@{}}
\toprule
\textbf{Category} & \textbf{Version} & \textbf{Prompt Template} \\ \midrule
\textbf{Baseline} & Original & Given the following \{text\}, rate its overall \{task\}. Respond using one of the following options: \{options\_string\}. Respond with an answer in English, without any explanation. \\ \cmidrule(l){2-3} 
 & Rephrased 1 & Analyze the \{task\} of the \{text\} provided below. Select the most appropriate rating from this list: \{options\_string\}. Give your response in English. Do not provide any context, reasoning, or additional text. \\ \cmidrule(l){2-3} 
 & Rephrased 2 & Your task is to provide a \{task\} rating for the following \{text\}. You must use one of these predefined labels: \{options\_string\}. Return only the English label as your final answer, skipping all introductory or explanatory remarks. \\ \midrule
\textbf{Demographic} & Original & Given the following text: \{text\}, how would a person of \{demographic\_attributes\} rate this \{task\}? Respond using one of the following options: \{options\_string\}. Respond with an answer in English, without any explanation. \\ \cmidrule(l){2-3} 
 & Rephrased 1 & Adopt the perspective of an individual with the following background: \{demographic\_attributes\}. Based on that viewpoint, evaluate the \{task\} of this text: \{text\}.  Choose from: \{options\_string\}. Response must be in English and contain only the selected option. \\ \cmidrule(l){2-3} 
 & Rephrased 2 & If a person identifying as \{demographic\_attributes\} were to read the text below, how would they likely rate its \{task\}? Choose from these given options: \{options\_string\}. Provide a one-word/phrase answer in English with no explanation. \\ \bottomrule
\end{tabular}
\caption{Baseline and with demographic Prompting Templates}
\label{tab:prompt-templates}
\end{table*}

\section{Detailed Model Results for Each Task}
\label{app:detail_model_results}

In this section, we provide the full performance breakdown for all five LLMs across the five experimental tasks. Tables \ref{tab:combined_toxicity_kappa_components} through \ref{tab:combined_emotion_kappa_components} present the results for the baseline, single-component, and multi-component configurations.

Specifically, Table \ref{tab:combined_toxicity_kappa_components} and Table \ref{tab:combined_sentiment_kappa_components} report the quadratic-weighted Cohen's $\kappa$ for the Toxicity and Sentiment tasks, respectively. For the Politeness and Offensiveness tasks, Cohen's $\kappa$ results are detailed in Table \ref{tab:combined_politeness_kappa_components} and Table \ref{tab:combined_offensiveness_kappa_components}. Table \ref{tab:combined_emotion_kappa_components} presents the classification accuracy for the Emotion task, as this dataset lacks natural ordinality. 

Furthermore, we provide comprehensive alignment metrics including Cohen's $\kappa$, Micro Accuracy, and Macro Accuracy for each task to facilitate a more granular analysis: Toxicity (Table \ref{tab:combined_toxicity_micro_components_top_half}), Sentiment (Table \ref{tab:combined_sentiment_micro_components_top_half}), Politeness (Table \ref{tab:combined_politeness_micro_components_top_half}), and Offensiveness (Table \ref{tab:combined_offensiveness_kappa_components}). In these tables, significant improvements or degradations relative to the no-demographic baseline are indicated based on a two-sided paired bootstrap test ($B=10,000$).

\begin{table*}[t]
\centering
\small
\resizebox{\textwidth}{!}{%
\begin{tabular}{r l | r  r  r  r  r}
\hline
& & \textbf{Gemma} & \textbf{Llama} & \textbf{Qwen} & \textbf{DeepSeek} & \textbf{Mistral} \\
\textbf{no.} & \textbf{demographic in the prompt} & $\kappa$ & $\kappa$ & $\kappa$ & $\kappa$ & $\kappa$ \\
\hline
\multicolumn{7}{l}{\emph{\textbf{Baseline, single-component and all components together configurations}}} \\
0 & baseline (no demographic)  & 0.230             & 0.275             & 0.157             & 0.100             & 0.276             \\
1 & age\_range                 & 0.213             & 0.226$^{\dagger}$ & 0.151             & 0.057$^{\dagger}$ & 0.237$^{\dagger}$ \\
1 & education                  & 0.235             & 0.224$^{\dagger}$ & 0.198$^{*}$       & 0.060$^{\dagger}$ & 0.234$^{\dagger}$ \\
1 & gender                     & 0.191$^{\dagger}$ & 0.249$^{\dagger}$ & 0.141             & 0.068$^{\dagger}$ & 0.263             \\
1 & is\_parent                 & 0.208$^{\dagger}$ & 0.235$^{\dagger}$ & 0.168             & 0.078$^{\dagger}$ & 0.242$^{\dagger}$ \\
1 & lgbtq\_status              & 0.236             & 0.264             & 0.246$^{*}$       & 0.055$^{\dagger}$ & 0.316$^{*}$       \\
1 & political\_affilation      & 0.234             & 0.251$^{\dagger}$ & 0.192$^{*}$       & 0.059$^{\dagger}$ & 0.299$^{*}$       \\
1 & race                       & 0.195$^{\dagger}$ & 0.289             & 0.172             & 0.064$^{\dagger}$ & 0.267             \\
1 & religion\_important        & 0.271$^{*}$       & 0.270             & 0.242$^{*}$       & 0.072$^{\dagger}$ & 0.287             \\
8 & all together               & 0.231             & 0.258             & 0.240$^{*}$       & 0.070$^{\dagger}$ & 0.271             \\
\hline
\multicolumn{7}{l}{\emph{\textbf{Top 5 configurations (no. of components $\geq$ 2 and not all together) exceeding baseline $\kappa$ for Gemma}}} \\
3 & education, is\_parent, religion\_important                              & 0.266$^{*}$ & -- & -- & -- & -- \\
2 & education, religion\_important                                          & 0.266$^{*}$ & -- & -- & -- & -- \\
3 & education, political\_affilation, religion\_important                   & 0.260$^{*}$ & -- & -- & -- & -- \\
3 & education, lgbtq\_status, religion\_important                           & 0.258$^{*}$ & -- & -- & -- & -- \\
2 & political\_affilation, religion\_important                              & 0.257$^{*}$ & -- & -- & -- & -- \\
\hline
\multicolumn{7}{l}{\emph{\textbf{Top 5 configurations (no. of components $\geq$ 2 and not all together) exceeding baseline $\kappa$ for Llama}}} \\
2 & lgbtq\_status, race                                                     & -- & 0.299$^{*}$ & -- & -- & -- \\
4 & is\_parent, lgbtq\_status, political\_affilation, race                  & -- & 0.290$^{*}$       & -- & -- & -- \\
3 & lgbtq\_status, political\_affilation, race                              & -- & 0.289       & -- & -- & -- \\
3 & education, gender, lgbtq\_status                                        & -- & 0.286       & -- & -- & -- \\
3 & is\_parent, lgbtq\_status, race                                         & -- & 0.285       & -- & -- & -- \\
\hline
\multicolumn{7}{l}{\emph{\textbf{Top 5 configurations (no. of components $\geq$ 2 and not all together) exceeding baseline $\kappa$ for Qwen}}} \\
3 & age\_range, lgbtq\_status, religion\_important                          & -- & -- & 0.294$^{*}$ & -- & -- \\
2 & lgbtq\_status, race                                                     & -- & -- & 0.289$^{*}$ & -- & -- \\
4 & gender, lgbtq\_status, political\_affilation, religion\_important       & -- & -- & 0.288$^{*}$ & -- & -- \\
3 & education, lgbtq\_status, race                                          & -- & -- & 0.286$^{*}$ & -- & -- \\
4 & age\_range, education, lgbtq\_status, race                              & -- & -- & 0.284$^{*}$ & -- & -- \\
\hline
\multicolumn{7}{l}{\emph{\textbf{Top 5 configurations (no. of components $\geq$ 2 and not all together) exceeding baseline $\kappa$ for DeepSeek}}} \\
-- & [No entries]                                                            & -- & -- & -- & -- & -- \\
\hline
\multicolumn{7}{l}{\emph{\textbf{Top 5 configurations (no. of components $\geq$ 2 and not all together) exceeding baseline $\kappa$ for Mistral}}} \\
2 & lgbtq\_status, political\_affiliation                                    & -- & -- & -- & -- & 0.323$^{*}$ \\
2 & lgbtq\_status, religion\_important                                      & -- & -- & -- & -- & 0.316$^{*}$ \\
3 & lgbtq\_status, political\_affilation, religion\_important               & -- & -- & -- & -- & 0.313$^{*}$ \\
3 & is\_parent, lgbtq\_status, political\_affiliation                        & -- & -- & -- & -- & 0.304$^{*}$ \\
4 & gender, lgbtq\_status, political\_affiliation, race                      & -- & -- & -- & -- & 0.303$^{*}$ \\
\hline
\end{tabular}%
}
\caption{\textbf{Cohen's $\kappa$ on the Toxicity dataset} for Gemma, Llama, Qwen, DeepSeek, and Mistral.
 Statistical significance is assessed via a two-sided paired bootstrap test ($B$\,=\,10{,}000 resamples);
we report the 95\% confidence interval on $\Delta\kappa = \kappa_{\text{config}} - \kappa_{\text{baseline}}$
and consider a result significant when the CI excludes zero ($p$\,<\,0.05). 
% Given the large sample size
% ($N$\,=\,16{,}000), we only interpret differences as practically meaningful when $|\Delta\kappa| \geq 0.02$
% to avoid over-interpreting trivially small effects.
$^{*}$~significant improvement; $^{\dagger}$~significant degradation relative to baseline.}
\label{tab:combined_toxicity_kappa_components}
\end{table*}

\begin{table*}[t]
\centering
\small
\resizebox{\textwidth}{!}{%
\begin{tabular}{r l | r  r  r  r  r}
\hline
& & \textbf{Gemma} & \textbf{Llama} & \textbf{Qwen} & \textbf{DeepSeek} & \textbf{Mistral} \\
\textbf{no.} & \textbf{demographic in the prompt} & $\kappa$ & $\kappa$ & $\kappa$ & $\kappa$ & $\kappa$ \\
\hline
\multicolumn{7}{l}{\emph{\textbf{Baseline, single-component and all components together configurations}}} \\
0 & baseline (no demographic) & 0.388                & 0.347                & 0.337                & 0.265                & 0.419                \\
1 & age                       & 0.355$^{\dagger}$    & 0.335                & 0.315$^{\dagger}$    & 0.241$^{\dagger}$    & 0.405                \\
1 & education                 & 0.354$^{\dagger}$    & 0.330                & 0.346                & 0.240$^{\dagger}$    & 0.415                \\
1 & employment\_status        & 0.339$^{\dagger}$    & 0.321$^{\dagger}$    & 0.310$^{\dagger}$    & 0.254                & 0.406                \\
1 & gender                    & 0.366$^{\dagger}$    & 0.350                & 0.349                & 0.262                & 0.414                \\
1 & hispanic\_latino          & 0.339$^{\dagger}$    & 0.335                & 0.342                & 0.262                & 0.405                \\
1 & income                    & 0.338$^{\dagger}$    & 0.326$^{\dagger}$    & 0.301$^{\dagger}$    & 0.251                & 0.409                \\
1 & living\_situation         & 0.352$^{\dagger}$    & 0.333                & 0.326                & 0.260                & 0.417                \\
1 & political\_id             & 0.300$^{\dagger}$    & 0.303$^{\dagger}$    & 0.324                & 0.239$^{\dagger}$    & 0.396$^{\dagger}$    \\
1 & race                      & 0.362$^{\dagger}$    & 0.343                & 0.345                & 0.246                & 0.410                \\
9 & all together              & 0.305$^{\dagger}$    & 0.281$^{\dagger}$    & 0.267$^{\dagger}$    & 0.227$^{\dagger}$    & 0.401                \\
\hline
\multicolumn{7}{l}{\emph{\textbf{Top 5 configurations (no. of components $\geq$ 2 and not all together) exceeding baseline $\kappa$ for Gemma, Llama, Mistral}}} \\
-- & [No entries]                                                              & -- & -- & -- & -- & -- \\
\hline
\multicolumn{7}{l}{\emph{\textbf{Top 5 configurations (no. of components $\geq$ 2 and not all together) exceeding baseline $\kappa$ for Qwen}}} \\
2 & education, hispanic\_latino                                                & -- & -- & 0.357$^{*}$  & -- & -- \\
2 & education, race                                                            & -- & -- & 0.354        & -- & -- \\
3 & education, gender, hispanic\_latino                                        & -- & -- & 0.353        & -- & -- \\
3 & employment\_status, gender, hispanic\_latino                               & -- & -- & 0.352        & -- & -- \\
3 & age, gender, hispanic\_latino                                              & -- & -- & 0.350        & -- & -- \\
\hline
\multicolumn{7}{l}{\emph{\textbf{Top 5 configurations (no. of components $\geq$ 2 and not all together) exceeding baseline $\kappa$ for DeepSeek}}} \\
5 & education, gender, income, living\_situation, race                         & -- & -- & -- & 0.291$^{*}$ & -- \\
5 & age, employment\_status, hispanic\_latino, income, race                    & -- & -- & -- & 0.283       & -- \\
5 & age, employment\_status, hispanic\_latino, living\_situation, race         & -- & -- & -- & 0.282       & -- \\
5 & employment\_status, gender, living\_situation, political\_id, race         & -- & -- & -- & 0.281       & -- \\
5 & age, gender, hispanic\_latino, living\_situation, race                     & -- & -- & -- & 0.278       & -- \\
\hline
\end{tabular}%
}
\caption{\textbf{Cohen's $\kappa$ on the Sentiment dataset} for Gemma, Llama, Qwen, DeepSeek, and Mistral. $^{*}$~significant improvement; $^{\dagger}$~significant degradation relative to baseline, calculated in the similar way as in \Cref{tab:combined_toxicity_kappa_components}.}
\label{tab:combined_sentiment_kappa_components}
\end{table*}

\begin{table*}[t]
\centering
\small
\resizebox{\textwidth}{!}{%
\begin{tabular}{r l | *{5}{>{\centering\arraybackslash}p{1.8cm}}}
\hline
& & \textbf{Gemma} & \textbf{Llama} & \textbf{Qwen} & \textbf{DeepSeek} & \textbf{Mistral} \\
\textbf{no.} & \textbf{demographic in the prompt} & $\kappa$ & $\kappa$ & $\kappa$ & $\kappa$ & $\kappa$ \\
\hline
\multicolumn{7}{l}{\emph{\textbf{Baseline, single-component and all components together configurations}}} \\
0 & baseline (no demographic) & 0.372                & 0.349                & 0.435                & 0.244                & 0.448                \\
1 & age                       & 0.376                & 0.253$^{\dagger}$    & 0.376$^{\dagger}$    & 0.109$^{\dagger}$    & 0.354$^{\dagger}$    \\
1 & education                 & 0.383                & 0.237$^{\dagger}$    & 0.361$^{\dagger}$    & 0.137$^{\dagger}$    & 0.349$^{\dagger}$    \\
1 & gender                    & 0.408$^{*}$          & 0.282$^{\dagger}$    & 0.425                & 0.105$^{\dagger}$    & 0.368$^{\dagger}$    \\
1 & occupation                & 0.411$^{*}$          & 0.267$^{\dagger}$    & 0.389$^{\dagger}$    & 0.117$^{\dagger}$    & 0.370$^{\dagger}$    \\
1 & race                      & 0.407$^{*}$          & 0.252$^{\dagger}$    & 0.409$^{\dagger}$    & 0.101$^{\dagger}$    & 0.353$^{\dagger}$    \\
5 & all together              & 0.393$^{*}$          & 0.239$^{\dagger}$    & 0.375$^{\dagger}$    & 0.119$^{\dagger}$    & 0.365$^{\dagger}$    \\
\hline
\multicolumn{7}{l}{\emph{\textbf{Top 5 configurations (no. of components $\geq$ 2 and not all together) exceeding baseline $\kappa$ for Gemma}}} \\
2 & education, occupation               & 0.406$^{*}$ & -- & -- & -- & -- \\
2 & gender, race                        & 0.404$^{*}$ & -- & -- & -- & -- \\
2 & gender, occupation                  & 0.401$^{*}$ & -- & -- & -- & -- \\
4 & education, gender, occupation, race & 0.399$^{*}$ & -- & -- & -- & -- \\
2 & education, race                     & 0.398$^{*}$ & -- & -- & -- & -- \\
\hline
\multicolumn{7}{l}{\emph{\textbf{Top 5 configurations (no. of components $\geq$ 2 and not all together) exceeding baseline $\kappa$ for Llama, Qwen, DeepSeek, and Mistral}}} \\
-- & [No entries]                       & -- & -- & -- & -- & -- \\
\hline
\end{tabular}%
}
\caption{\textbf{Cohen's $\kappa$ on the Politeness dataset} for Gemma, Llama, Qwen, DeepSeek, and Mistral. $^{*}$~significant improvement; $^{\dagger}$~significant degradation relative to baseline, calculated in the similar way as in \Cref{tab:combined_toxicity_kappa_components}.}
\label{tab:combined_politeness_kappa_components}
\end{table*}

\begin{table*}[t]
\centering
\small
\resizebox{\textwidth}{!}{%
\begin{tabularx}{\textwidth}{r l | *{5}{>{\centering\arraybackslash}X}}
\hline
& & \textbf{Gemma} & \textbf{Llama} & \textbf{Qwen} & \textbf{DeepSeek} & \textbf{Mistral} \\
\textbf{no.} & \textbf{demographic in the prompt} & $\kappa$ & $\kappa$ & $\kappa$ & $\kappa$ & $\kappa$ \\
\hline
\multicolumn{7}{l}{\emph{\textbf{Baseline, single-component and all components together configurations}}} \\
0 & baseline (no demographic) & 0.248                & 0.186                & 0.205                & 0.101                & 0.243                \\
1 & age                       & 0.152$^{\dagger}$    & 0.131$^{\dagger}$    & 0.158$^{\dagger}$    & 0.094                & 0.179$^{\dagger}$    \\
1 & education                 & 0.182$^{\dagger}$    & 0.137$^{\dagger}$    & 0.198                & 0.084                & 0.223$^{\dagger}$    \\
1 & gender                    & 0.160$^{\dagger}$    & 0.128$^{\dagger}$    & 0.209                & 0.088                & 0.219$^{\dagger}$    \\
1 & occupation                & 0.191$^{\dagger}$    & 0.146$^{\dagger}$    & 0.204                & 0.078$^{\dagger}$    & 0.227                \\
1 & race                      & 0.167$^{\dagger}$    & 0.139$^{\dagger}$    & 0.256$^{*}$          & 0.085                & 0.234                \\
5 & all together              & 0.175$^{\dagger}$    & 0.117$^{\dagger}$    & 0.205                & 0.087                & 0.223$^{\dagger}$    \\
\hline
\multicolumn{7}{l}{\emph{\textbf{Top 5 configurations (no. of components $\geq$ 2 and not all together) exceeding baseline $\kappa$ for Gemma, Llama, and DeepSeek}}} \\
-- & [No entries]              & -- & -- & -- & -- & -- \\
\hline
\multicolumn{7}{l}{\emph{\textbf{Top 5 configurations (no. of components $\geq$ 2 and not all together) exceeding baseline $\kappa$ for Qwen}}} \\
2 & gender, race                & -- & -- & 0.244$^{*}$ & -- & -- \\
2 & occupation, race            & -- & -- & 0.239$^{*}$ & -- & -- \\
2 & education, race             & -- & -- & 0.234$^{*}$ & -- & -- \\
3 & education, occupation, race & -- & -- & 0.232$^{*}$ & -- & -- \\
3 & gender, occupation, race    & -- & -- & 0.220       & -- & -- \\
\hline
\multicolumn{7}{l}{\emph{\textbf{Top 5 configurations (no. of components $\geq$ 2 and not all together) exceeding baseline $\kappa$ for Mistral}}} \\
3 & education, occupation, race & -- & -- & -- & -- & 0.246 \\
2 & occupation, race            & -- & -- & -- & -- & 0.244 \\
\hline
\end{tabularx}%
}
\caption{\textbf{Cohen's $\kappa$ on the Offensiveness dataset} for Gemma, Llama, Qwen, DeepSeek, and Mistral. $^{*}$~significant improvement; $^{\dagger}$~significant degradation relative to baseline, calculated in the similar way as in \Cref{tab:combined_toxicity_kappa_components}.}
\label{tab:combined_offensiveness_kappa_components}
\end{table*}

\begin{table*}[t]
\centering
\small
\resizebox{\textwidth}{!}{%
\begin{tabular}{r l | r  r  r  r  r}
\hline
& & \textbf{Gemma} & \textbf{Llama} & \textbf{Qwen} & \textbf{DeepSeek} & \textbf{Mistral} \\
\textbf{no.} & \textbf{demographic in the prompt} & Accuracy & Accuracy & Accuracy & Accuracy & Accuracy \\
\hline
\multicolumn{7}{l}{\emph{\textbf{Baseline, single-component and all components together configurations}}} \\
0 & baseline (no demographic) & 0.7256 & 0.5601 & 0.6032 & 0.3566 & 0.4829 \\
1 & age & 0.6871 & 0.5548 & 0.6407 & 0.3177 & 0.6114 \\
1 & country & 0.4741 & 0.4265 & 0.6250 & 0.3248 & 0.5710 \\
1 & father\_occupation & 0.4633 & 0.4567 & 0.5911 & 0.2986 & 0.5911 \\
1 & field\_of\_study & 0.4176 & 0.4720 & 0.6081 & 0.3099 & 0.5817 \\
1 & gender & 0.6538 & 0.5348 & 0.6114 & 0.3174 & 0.6066 \\
1 & mother\_occupation & 0.5229 & 0.4969 & 0.5768 & 0.2876 & 0.5713 \\
1 & religion & 0.5676 & 0.4682 & 0.5791 & 0.2962 & 0.5620 \\
7 & all together & 0.6255 & 0.5537 & 0.6234 & 0.3006 & 0.6008 \\
\hline
\multicolumn{7}{l}{\emph{\textbf{Top 5 configurations (no. of components $\geq$ 2 and not all together) exceeding baseline Accuracy for Gemma, DeepSeek}}} \\
-- & [No entries] & -- & -- & -- & -- & -- \\
\hline
\multicolumn{7}{l}{\emph{\textbf{Top 5 configurations (no. of components $\geq$ 2 and not all together) exceeding baseline Accuracy for Llama}}} \\
4 & age, father\_occupation, gender, mother\_occupation & -- & 0.5715 & -- & -- & -- \\
5 & age, country, father\_occupation, gender, mother\_occupation & -- & 0.5715 & -- & -- & -- \\
6 & age, country, father\_occupation, field\_of\_study, gender, mother\_occupation & -- & 0.5708 & -- & -- & -- \\
3 & age, gender, mother\_occupation & -- & 0.5693 & -- & -- & -- \\
4 & age, country, father\_occupation, mother\_occupation & -- & 0.5691 & -- & -- & -- \\
\hline
\multicolumn{7}{l}{\emph{\textbf{Top 5 configurations (no. of components $\geq$ 2 and not all together) exceeding baseline Accuracy for Qwen}}} \\
5 & age, country, father\_occupation, gender, mother\_occupation & -- & -- & 0.6301 & -- & -- \\
2 & age, country & -- & -- & 0.6295 & -- & -- \\
5 & age, father\_occupation, field\_of\_study, gender, mother\_occupation & -- & -- & 0.6285 & -- & -- \\
4 & age, country, father\_occupation, mother\_occupation & -- & -- & 0.6285 & -- & -- \\
3 & age, father\_occupation, gender & -- & -- & 0.6277 & -- & -- \\
\hline
% \multicolumn{7}{l}{\emph{\textbf{Top 5 configurations (no. of components $\geq$ 2 and not all together) exceeding baseline Accuracy for DeepSeek}}} \\
% -- & [No entries] & -- & -- & -- & -- & -- \\
% \hline
\multicolumn{7}{l}{\emph{\textbf{Top 5 configurations (no. of components $\geq$ 2 and not all together) exceeding baseline Accuracy for Mistral}}} \\
6 & country, father\_occupation, field\_of\_study, gender, mother\_occupation, religion & -- & -- & -- & -- & 0.5998 \\
5 & father\_occupation, field\_of\_study, gender, mother\_occupation, religion & -- & -- & -- & -- & 0.5995 \\
4 & father\_occupation, gender, mother\_occupation, religion & -- & -- & -- & -- & 0.5969 \\
3 & father\_occupation, gender, religion & -- & -- & -- & -- & 0.5923 \\
2 & gender, religion & -- & -- & -- & -- & 0.5872 \\
\hline
\end{tabular}%
}
\caption{\textbf{Accuracy on the Emotion dataset} for Gemma, Llama, Qwen, DeepSeek, and Mistral.}
\label{tab:combined_emotion_kappa_components}
\end{table*}

\begin{table*}[t]
\centering
\small
\resizebox{\textwidth}{!}{%
\begin{tabular}{r l | r r r | r r r | r r r | r r r | r r r}
\hline
& & \multicolumn{3}{c|}{\textbf{Gemma}} & \multicolumn{3}{c|}{\textbf{Llama}} & \multicolumn{3}{c|}{\textbf{Qwen}} & \multicolumn{3}{c|}{\textbf{DeepSeek}} & \multicolumn{3}{c}{\textbf{Mistral}} \\
\textbf{no.} & \textbf{demographic in the prompt} & $\kappa$ & Micro & Macro & $\kappa$ & Micro & Macro & $\kappa$ & Micro & Macro & $\kappa$ & Micro & Macro & $\kappa$ & Micro & Macro \\
\hline
\multicolumn{17}{l}{\emph{\textbf{Baseline, single-component, and all components together configurations}}} \\
0 & baseline (no demographic) & 0.230 & 0.222 & 0.235 & 0.275 & 0.300 & 0.268 & 0.157 & 0.214 & 0.243 & 0.100 & 0.286 & 0.228 & 0.276 & 0.244 & 0.271 \\
1 & age\_range                 & 0.213 & 0.139 & 0.236 & 0.226 & 0.287 & 0.256 & 0.151 & 0.171 & 0.224 & 0.057 & 0.196 & 0.210 & 0.237 & 0.197 & 0.263 \\
1 & education                 & 0.235 & 0.161 & 0.245 & 0.224 & 0.305 & 0.256 & 0.198 & 0.226 & 0.253 & 0.060 & 0.204 & 0.212 & 0.234 & 0.201 & 0.260 \\
1 & gender                    & 0.191 & 0.137 & 0.235 & 0.249 & 0.322 & 0.262 & 0.141 & 0.178 & 0.233 & 0.068 & 0.203 & 0.212 & 0.263 & 0.230 & 0.271 \\
1 & is\_parent                & 0.208 & 0.155 & 0.242 & 0.235 & 0.303 & 0.263 & 0.168 & 0.228 & 0.248 & 0.078 & 0.206 & 0.214 & 0.242 & 0.211 & 0.263 \\
1 & lgbtq\_status             & 0.236 & 0.192 & 0.255 & 0.264 & 0.456 & 0.262 & 0.246 & 0.393 & 0.263 & 0.055 & 0.200 & 0.208 & 0.316 & 0.334 & 0.299 \\
1 & political\_affilation     & 0.234 & 0.170 & 0.251 & 0.251 & 0.327 & 0.267 & 0.192 & 0.222 & 0.242 & 0.059 & 0.206 & 0.213 & 0.299 & 0.278 & 0.293 \\
1 & race                      & 0.195 & 0.163 & 0.251 & 0.289 & 0.397 & 0.280 & 0.172 & 0.222 & 0.244 & 0.064 & 0.206 & 0.208 & 0.267 & 0.249 & 0.274 \\
1 & religion\_important       & 0.271 & 0.213 & 0.268 & 0.270 & 0.354 & 0.270 & 0.242 & 0.295 & 0.266 & 0.072 & 0.194 & 0.209 & 0.287 & 0.244 & 0.275 \\
8 & all components together              & 0.231 & 0.167 & 0.251 & 0.258 & 0.319 & 0.270 & 0.240 & 0.307 & 0.273 & 0.070 & 0.199 & 0.215 & 0.271 & 0.237 & 0.274 \\
\hline
\end{tabular}%
}
\caption{\textbf{Cohen $\kappa$, Micro Accuracy, and Macro Accuracy for Toxicity Dataset}. This table lists the baseline, all single-component, and all components together configurations.}
\label{tab:combined_toxicity_micro_components_top_half}
\end{table*}

\begin{table*}[t]
\centering
\small
\resizebox{\textwidth}{!}{%
\begin{tabular}{r l | r r r | r r r | r r r | r r r | r r r}
\hline
& & \multicolumn{3}{c|}{\textbf{Gemma}} & \multicolumn{3}{c|}{\textbf{Llama}} & \multicolumn{3}{c|}{\textbf{Qwen}} & \multicolumn{3}{c|}{\textbf{DeepSeek}} & \multicolumn{3}{c}{\textbf{Mistral}} \\
\textbf{no.} & \textbf{demographic in the prompt} & $\kappa$ & Micro & Macro & $\kappa$ & Micro & Macro & $\kappa$ & Micro & Macro & $\kappa$ & Micro & Macro & $\kappa$ & Micro & Macro \\
\hline
\multicolumn{17}{l}{\emph{\textbf{Baseline and single-component configurations (shared across models)}}} \\
0 & baseline (no demographic) & 0.388 & 0.269 & 0.360 & 0.347 & 0.237 & 0.343 & 0.337 & 0.279 & 0.342 & 0.265 & 0.290 & 0.290 & 0.419 & 0.318 & 0.358 \\
1 & age                       & 0.355 & 0.270 & 0.330 & 0.335 & 0.253 & 0.327 & 0.315 & 0.233 & 0.336 & 0.241 & 0.267 & 0.272 & 0.405 & 0.301 & 0.359 \\
1 & education                 & 0.354 & 0.278 & 0.329 & 0.330 & 0.274 & 0.311 & 0.346 & 0.280 & 0.353 & 0.240 & 0.297 & 0.276 & 0.415 & 0.321 & 0.353 \\
1 & employment\_status        & 0.339 & 0.267 & 0.322 & 0.321 & 0.249 & 0.313 & 0.310 & 0.255 & 0.329 & 0.254 & 0.287 & 0.270 & 0.406 & 0.316 & 0.358 \\
1 & gender                    & 0.366 & 0.262 & 0.344 & 0.350 & 0.256 & 0.333 & 0.349 & 0.269 & 0.350 & 0.262 & 0.283 & 0.283 & 0.414 & 0.313 & 0.347 \\
1 & hispanic\_latino          & 0.339 & 0.274 & 0.323 & 0.335 & 0.265 & 0.318 & 0.342 & 0.312 & 0.341 & 0.262 & 0.303 & 0.278 & 0.405 & 0.334 & 0.348 \\
1 & income                    & 0.338 & 0.276 & 0.318 & 0.326 & 0.275 & 0.324 & 0.301 & 0.273 & 0.327 & 0.251 & 0.290 & 0.272 & 0.409 & 0.332 & 0.350 \\
1 & living\_situation         & 0.352 & 0.267 & 0.332 & 0.333 & 0.257 & 0.317 & 0.326 & 0.266 & 0.347 & 0.260 & 0.288 & 0.286 & 0.417 & 0.322 & 0.360 \\
1 & political\_id             & 0.300 & 0.296 & 0.307 & 0.303 & 0.263 & 0.298 & 0.324 & 0.290 & 0.342 & 0.239 & 0.312 & 0.266 & 0.396 & 0.341 & 0.340 \\
1 & race                      & 0.362 & 0.285 & 0.337 & 0.343 & 0.269 & 0.319 & 0.345 & 0.304 & 0.346 & 0.246 & 0.301 & 0.282 & 0.410 & 0.341 & 0.348 \\
9 & all together              & 0.305 & 0.289 & 0.312 & 0.281 & 0.262 & 0.300 & 0.267 & 0.252 & 0.330 & 0.227 & 0.278 & 0.266 & 0.401 & 0.331 & 0.346 \\
\hline
\end{tabular}%
}
\caption{\textbf{Cohen $\kappa$, Micro Accuracy, and Macro Accuracy for Sentiment Dataset}. This table lists the baseline, all single-component, and all components together configurations.}
\label{tab:combined_sentiment_micro_components_top_half}
\end{table*}

\begin{table*}[t]
\centering
\small
\resizebox{\textwidth}{!}{%
\begin{tabular}{r l | r r r | r r r | r r r | r r r | r r r}
\hline
& & \multicolumn{3}{c|}{\textbf{Gemma}} & \multicolumn{3}{c|}{\textbf{Llama}} & \multicolumn{3}{c|}{\textbf{Qwen}} & \multicolumn{3}{c|}{\textbf{DeepSeek}} & \multicolumn{3}{c}{\textbf{Mistral}} \\
\textbf{no.} & \textbf{demographic in the prompt} & $\kappa$ & Micro & Macro & $\kappa$ & Micro & Macro & $\kappa$ & Micro & Macro & $\kappa$ & Micro & Macro & $\kappa$ & Micro & Macro \\
\hline
\multicolumn{17}{l}{\emph{\textbf{Baseline and single-component configurations (shared across models)}}} \\
0 & baseline (no demographic) & 0.372 & 0.302 & 0.317 & 0.349 & 0.263 & 0.327 & 0.435 & 0.295 & 0.337 & 0.244 & 0.276 & 0.266 & 0.448 & 0.314 & 0.329 \\
1 & age                       & 0.376 & 0.307 & 0.293 & 0.253 & 0.298 & 0.271 & 0.376 & 0.282 & 0.315 & 0.109 & 0.265 & 0.226 & 0.354 & 0.284 & 0.291 \\
1 & education                 & 0.383 & 0.307 & 0.300 & 0.237 & 0.291 & 0.269 & 0.361 & 0.277 & 0.314 & 0.137 & 0.271 & 0.235 & 0.349 & 0.285 & 0.292 \\
1 & gender                    & 0.408 & 0.315 & 0.315 & 0.282 & 0.305 & 0.283 & 0.425 & 0.303 & 0.334 & 0.105 & 0.262 & 0.224 & 0.368 & 0.287 & 0.298 \\
1 & occupation                & 0.411 & 0.307 & 0.314 & 0.267 & 0.289 & 0.275 & 0.389 & 0.285 & 0.322 & 0.117 & 0.266 & 0.227 & 0.370 & 0.287 & 0.297 \\
1 & race                      & 0.407 & 0.308 & 0.314 & 0.252 & 0.305 & 0.272 & 0.409 & 0.297 & 0.330 & 0.101 & 0.256 & 0.223 & 0.353 & 0.285 & 0.294 \\
5 & all together              & 0.393 & 0.307 & 0.300 & 0.239 & 0.293 & 0.267 & 0.375 & 0.279 & 0.319 & 0.119 & 0.262 & 0.227 & 0.365 & 0.283 & 0.292 \\
\hline
\end{tabular}%
}
\caption{\textbf{Cohen $\kappa$, Micro Accuracy, and Macro Accuracy for Politeness Dataset}. This table lists the baseline, all single-component, and all components together configurations.}
\label{tab:combined_politeness_micro_components_top_half}
\end{table*}

\begin{table*}[t]
\centering
\small
\resizebox{\textwidth}{!}{%
\begin{tabular}{r l | r r r | r r r | r r r | r r r | r r r}
\hline
& & \multicolumn{3}{c|}{\textbf{Gemma}} & \multicolumn{3}{c|}{\textbf{Llama}} & \multicolumn{3}{c|}{\textbf{Qwen}} & \multicolumn{3}{c|}{\textbf{DeepSeek}} & \multicolumn{3}{c}{\textbf{Mistral}} \\
\textbf{no.} & \textbf{demographic in the prompt} & $\kappa$ & Micro & Macro & $\kappa$ & Micro & Macro & $\kappa$ & Micro & Macro & $\kappa$ & Micro & Macro & $\kappa$ & Micro & Macro \\
\hline
\multicolumn{17}{l}{\emph{\textbf{Baseline and single-component configurations (shared across models)}}} \\
0 & baseline (no demographic) & 0.248 & 0.218 & 0.288 & 0.186 & 0.260 & 0.282 & 0.205 & 0.291 & 0.288 & 0.101 & 0.313 & 0.222 & 0.284 & 0.329 & 0.308 \\
1 & age                       & 0.152 & 0.118 & 0.252 & 0.131 & 0.139 & 0.253 & 0.158 & 0.232 & 0.277 & 0.094 & 0.261 & 0.226 & 0.179 & 0.209 & 0.280 \\
1 & education                 & 0.182 & 0.137 & 0.265 & 0.137 & 0.140 & 0.253 & 0.198 & 0.271 & 0.287 & 0.084 & 0.262 & 0.221 & 0.223 & 0.250 & 0.289 \\
1 & gender                    & 0.160 & 0.123 & 0.259 & 0.128 & 0.145 & 0.253 & 0.209 & 0.311 & 0.291 & 0.088 & 0.273 & 0.227 & 0.219 & 0.246 & 0.285 \\
1 & occupation                & 0.191 & 0.138 & 0.258 & 0.146 & 0.152 & 0.250 & 0.204 & 0.295 & 0.288 & 0.078 & 0.271 & 0.233 & 0.227 & 0.261 & 0.292 \\
1 & race                      & 0.167 & 0.138 & 0.270 & 0.139 & 0.179 & 0.260 & 0.256 & 0.384 & 0.308 & 0.085 & 0.280 & 0.225 & 0.234 & 0.263 & 0.300 \\
5 & all together              & 0.175 & 0.134 & 0.261 & 0.117 & 0.125 & 0.247 & 0.205 & 0.299 & 0.297 & 0.087 & 0.274 & 0.223 & 0.223 & 0.256 & 0.294 \\
\hline
\end{tabular}%
}
\caption{\textbf{Cohen $\kappa$, Micro Accuracy, and Macro Accuracy for Offensiveness Dataset}. This table lists the baseline, all single-component, and all components together configurations.}
\label{tab:combined_offensiveness_micro_components_top_half}
\end{table*}

\section{SHAP (SHapley Additive exPlanations)}
\label{app:shap}

To interpret the model's decision-making process and evaluate the influence of demographic attributes on predictions, we utilized SHAP (SHapley Additive exPlanations). After training a \textit{Logistic Regression} model on One-Hot Encoded demographic data, we applied a LinearExplainer to compute the SHAP values for each instance. To assess feature importance at the categorical level, we aggregated the absolute SHAP values of the encoded binary features back to their parent demographic groups (e.g., combining gender Male and gender Female into gender). This approach allowed us to extract both global feature importance rankings and local, instance-level explanations, revealing the magnitude and directionality of how specific demographic groups bias the model toward or away from the original label.

\begin{table*}[ht]
\centering
\small
\resizebox{\textwidth}{!}{
\begin{tabular}{lc | lc | lc | lc | lc}
\hline
\multicolumn{2}{c|}{\textbf{Toxicity}} & \multicolumn{2}{c|}{\textbf{Sentiment}} & \multicolumn{2}{c|}{\textbf{Politeness}} & \multicolumn{2}{c|}{\textbf{Offensiveness}} & \multicolumn{2}{c}{\textbf{Emotion}} \\ \hline
\textbf{Attribute} & \textbf{SHAP} & \textbf{Attribute} & \textbf{SHAP} & \textbf{Attribute} & \textbf{SHAP} & \textbf{Attribute} & \textbf{SHAP} & \textbf{Attribute} & \textbf{SHAP} \\ \hline
Religion Importance & 0.1889 & Living situation & 0.1889 & Race & 0.2255 & Age & 0.3661 & Country & 0.2151 \\
Education & 0.1495 & Household income & 0.1459 & Age & 0.2200 & Occupation & 0.2165 & Age & 0.1731 \\
Is Parent & 0.1272 & Political identification & 0.1456 & Occupation & 0.2186 & Race & 0.1568 & Mother's Occupation & 0.1535 \\
Age Range & 0.1254 & Gender & 0.1067 & Education & 0.1799 & Education & 0.1546 & Religion & 0.1421 \\
Gender & 0.1250 & Age & 0.1039 & Gender & 0.1560 & Gender & 0.1060 & Father's Occupation & 0.1267 \\
Race & 0.1063 & Education & 0.0980 & & & & & Field of Study & 0.1047 \\
LGBTQ Status & 0.0997 & Employment status & 0.0833 & & & & & Sex (Gender) & 0.0848 \\
Political Affiliation & 0.0779 & Race & 0.0825 & & & & & & \\
& & Hispanic/Latino & 0.0452 & & & & & & \\ \hline
\end{tabular}
}
\caption{SHAP Feature Importance Across All Tasks}
\label{tab:shap_all_tasks}
\end{table*}

\begin{table*}[ht]
\centering
\small
\resizebox{\textwidth}{!}{
\begin{tabular}{lc | lc | lc | lc | lc}
\hline
\multicolumn{2}{c|}{\textbf{Toxicity}} & \multicolumn{2}{c|}{\textbf{Sentiment}} & \multicolumn{2}{c|}{\textbf{Politeness}} & \multicolumn{2}{c|}{\textbf{Offensiveness}} & \multicolumn{2}{c}{\textbf{Emotion}} \\ \hline
\textbf{Attribute} & \textbf{LSVC $\kappa$} & \textbf{Attribute} & \textbf{LSVC $\kappa$} & \textbf{Attribute} & \textbf{LSVC $\kappa$} & \textbf{Attribute} & \textbf{LSVC $\kappa$} & \textbf{Attribute} & \textbf{LSVC Acc} \\ \hline
Education & 0.2007 & Employment status & 0.2048 & Gender & 0.5067 & Age & 0.3164 & Country & 0.5769 \\
LGBTQ status & 0.1999 & Gender & 0.2027 & Race & 0.4993 & Education & 0.3098 & Field of study & 0.5587 \\
Race & 0.1857 & Race & 0.2019 & Education & 0.4845 & Gender & 0.3055 & Religion & 0.5463 \\
Religion important & 0.1773 & Hispanic/Latino & 0.2005 & Occupation & 0.4724 & Race & 0.3023 & Age & 0.5385 \\
Is parent & 0.1764 & Education & 0.1947 & Age & 0.4507 & Occupation & 0.2971 & Father occupation & 0.5385 \\
Age range & 0.1739 & Income & 0.1946 & & & & & Mother occupation & 0.5372 \\
Political affiliation & 0.1608 & Age & 0.1873 & & & & & Gender & 0.5339 \\
Gender & 0.1607 & Living situation & 0.1755 & & & & & & \\
& & Political id & 0.1614 & & & & & & \\ \hline
\end{tabular}
}
\caption{LSVC Performance Across All Tasks}
\label{tab:lsvc_performance_all}
\end{table*}

\section{Word–Demographic Interaction Features and Spearman Rank Correlation Analysis}
\label{app:word_demographic_interaction}

\subsection{Measuring Annotator Demographic Influence via Learned Interaction Features}

To quantify how annotator demographic characteristics shape the target label at the
lexical level, we trained a \textit{Linear Support Vector Classifier (LinearSVC)} with explicitly
constructed \textit{word–demographic interaction features}. The motivation for this design is to
move beyond post-hoc analysis: rather than training a model on text and demographics
separately and then examining correlations after the fact, we encode the relationship
directly into the feature space so that the model learns differential word weights during
training.

\subsubsection{Feature Construction}

For each comment in the dataset, we first constructed two independent feature
representations. \textbf{Text features were extracted using TF-IDF vectorization} (maximum
vocabulary of 30,000 unigrams, minimum document frequency of 2), producing a sparse
matrix $\mathbf{X}_{\text{words}}$ of shape $(n_{\text{samples}} \times n_{\text{text\_features}})$. 
Annotator demographic attributes such as race, gender, religion, education, LGBTQ status, 
political affiliation, parental status, and age range were encoded using one-hot 
encoding, producing a sparse matrix $\mathbf{X}_{\text{demo}}$ of shape 
$(n_{\text{samples}} \times n_{\text{dem\_categories}})$, where each column corresponds 
to one unique value within a demographic attribute (e.g., \texttt{gender\_Female}, 
\texttt{gender\_Male}, \texttt{gender\_Nonbinary}).

The core of our approach is the explicit construction of word–demographic interaction
features before model fitting. For each demographic indicator column $d_j$ in 
$\mathbf{X}_{\text{demo}}$, we compute the element-wise product of the entire TF-IDF 
word matrix with that column:

\begin{equation}
\mathbf{X}_{\text{inter},j} = \mathbf{X}_{\text{words}} \odot d_j
\end{equation}

\noindent where $\odot$ denotes column-wise multiplication (each word's TF-IDF value is 
multiplied by the annotator's demographic indicator for category $j$). This produces one 
interaction block per demographic category, yielding an interaction matrix of shape
$(n_{\text{samples}} \times n_{\text{text\_features}} \times n_{\text{dem\_categories}})$. 
The final feature matrix passed to the classifier is the horizontal concatenation:

% \begin{equation}
% \mathbf{X}_{\text{full}} = \left[ \mathbf{X}_{\text{words}} \mid \mathbf{X}_{\text{demo}} \mid 
% \mathbf{X}_{\text{inter},1} \mid \mathbf{X}_{\text{inter},2} \mid \cdots \mid 
% \mathbf{X}_{\text{inter},k} \right]
% \end{equation}

\begin{equation}
\begin{split}
\mathbf{X}_{\text{full}} = [ \mathbf{X}_{\text{words}} \mid \mathbf{X}_{\text{demo}} \mid \mathbf{X}_{\text{inter},1} \\
\mid \mathbf{X}_{\text{inter},2} \mid \cdots \mid \mathbf{X}_{\text{inter},k} ]
\end{split}
\end{equation}

This design means the classifier learns a separate weight for every (word, demographic
category) pair for example, $w[\texttt{gender\_Female} \times \text{``threat''}]$ is 
learned independently from $w[\texttt{gender\_Male} \times \text{``threat''}]$. 
Crucially, these weights are estimated jointly during a single model fit rather than 
derived post-hoc, so they reflect the interaction as the model actually uses it in 
prediction. We deliberately exclude word–word interactions to avoid feature explosion 
while preserving full interpretability of the learned weights. Each model was trained 
separately per demographic attribute on an 80/20 stratified split, with a text-only 
baseline trained under identical conditions for comparison.

\subsubsection{Extracting Demographic-Specific Word Weights}

After fitting, we extracted the learned interaction weights from the model's coefficient
vector. For each demographic attribute (e.g., gender), we identified the top-200
most influential words by selecting the words with the highest maximum absolute
interaction weight across all demographic categories within that attribute. This
max-pooling selection ensures that the vocabulary of 200 words is not dominated by a
single category, and that the resulting word set is representative of lexical signals
that matter for at least one group.

For each word in this vocabulary, we then retrieved the learned interaction weight for
every demographic category within the attribute---regardless of whether that
(category, word) pair ranked in the top 200. This produces a wide-format weight table
$\mathbf{W}$ of shape $(n_{\text{words}} \times n_{\text{categories}})$, where entry 
$W[\text{word}, \text{category}]$ is the model's learned weight for how strongly that 
word pushes toward a toxic classification when the annotator belongs to that demographic 
category. A weight of zero indicates that the word was either absent from the training 
vocabulary for that category or had no discriminative value in that context.

\subsubsection{Spearman Rank Correlation Across Demographic Categories}

To quantify the degree to which two demographic categories produce similar lexical
toxicity patterns, we computed Spearman rank correlation coefficients between the weight
vectors of every pair of categories within each demographic attribute. For two categories
$c_i$ and $c_j$, the Spearman correlation $\rho(c_i, c_j)$ measures the rank-order 
agreement between the vectors $\mathbf{W}[:, c_i]$ and $\mathbf{W}[:, c_j]$ across the 
shared 200-word vocabulary.

% We chose Spearman over Pearson correlation for two reasons. First, the learned SVC
% weights are not normally distributed---they tend to be heavy-tailed with a small number
% of highly influential words. Second, we are primarily interested in whether two
% categories agree on the \textit{ordering} of word importance rather than the exact 
% magnitude of their weights, since magnitudes are not directly comparable across models 
% trained with different feature distributions.

A high positive $\rho$ indicates that the two demographic groups assign similar relative
importance to the same words when judging toxicity. A negative $\rho$ indicates that 
words strongly associated with toxicity for one group tend to be strongly associated with
non-toxicity for the other a pattern that directly reflects divergent annotation
behavior. A near-zero $\rho$ indicates no systematic lexical alignment between the two 
groups.

For each demographic attribute with $k$ unique categories, this procedure yields 
$\binom{k}{2}$ pairwise correlations. To summarize the overall degree of within-attribute 
divergence as a single comparable value, we aggregated pairwise correlations using the 
Fisher $z$-transformation:

\begin{align}
z_i &= \text{arctanh}(\rho_i) \\
\bar{z} &= \frac{1}{n} \sum_{i=1}^{n} z_i \\
\bar{\rho} &= \tanh(\bar{z})
\end{align}

\noindent where the sum runs over all $\binom{k}{2}$ pairs. Fisher $z$-averaging is 
preferred over the arithmetic mean of $\rho$ values because $\rho$ is bounded in 
$[-1, 1]$ and compressed near the extremes; averaging in $z$-space and back-transforming 
avoids this distortion and produces a more reliable estimate of the central tendency 
across pairs. 

\subsubsection{Reporting Strategy}

Because the number of pairwise comparisons grows quadratically with the number of
demographic categories (reaching $\binom{48}{2} = 1{,}128$ pairs for race alone) it is 
neither practical nor informative to report all individual correlations in the main text. 
We therefore adopt a two-level reporting structure. In Tables~\ref{tab:spearman_summary_toxicity}--\ref{tab:spearman_summary_emotion}, we
report the Fisher-averaged $\bar{\rho}$ as a compact cross-attribute divergence score 
alongside the single most divergent pair (highest $|\rho|$) within each attribute. This 
dominant pair captures the most salient signal within the attribute and flags cases where 
the summary $\bar{\rho}$ may be masking a strong but localized divergence. For gender, 
for example, the Fisher-averaged $\bar{\rho} = -0.087$ (non-significant at the aggregate 
level), but this summary conceals a strongly negative Female/Male correlation of 
$\rho = -0.629$ ($p < 0.001$), which indicates that words driving toxicity predictions 
for Female annotators systematically suppress them for Male annotators and vice versa.

\begin{table*}[htbp]
\centering
\small
\resizebox{\textwidth}{!}{%
\begin{tabular}{lrlr}
\toprule
\textbf{Demographic} & \textbf{Fisher $\bar{\rho}$} & \textbf{Dominant Pair} & \textbf{Dominant $\rho$} \\
\midrule
race & 0.2503*** & Black or African American, Hispanic vs White, Black or African American, American Indian or Alaska Native & 0.521*** \\
gender & -0.0869*** & Female vs Male & -0.629*** \\
religion\_important & -0.0009*** & important\_Not important vs important\_Very important & -0.365*** \\
education & 0.1589*** & Associate degree in college (2-year) vs Professional degree (JD, MD) & 0.489*** \\
lgbtq\_status & 0.2082*** & status\_Homosexual vs status\_Other & 0.409*** \\
political\_affilation & 0.0019*** & affilation\_Independent vs affilation\_Liberal & -0.399*** \\
is\_parent & -0.2028*** & parent\_No vs parent\_Yes & -0.636*** \\
age\_range & 0.1245*** & range\_25 - 34 vs range\_35 - 44 & -0.489*** \\
\bottomrule
\end{tabular}%
}
\caption{Summary of Spearman Correlations by Demographic Category for \textbf{Toxicity dataset}. {\small \textit{Note:} *** $p<.001$, ** $p<.01$, * $p<.05$, ns = not significant.}}
\label{tab:spearman_summary_toxicity}
\end{table*}

\begin{table*}[htbp]
\centering
\small
\resizebox{\textwidth}{!}{%
\begin{tabular}{lrlr}
\toprule
\textbf{Demographic} & \textbf{Fisher $\bar{\rho}$} & \textbf{Dominant Pair} & \textbf{Dominant $\rho$} \\
\midrule
age & 0.0319*** & 50-59 vs 60-69 & -0.475*** \\
race & 0.2074*** & Black or African American vs Other & 0.452*** \\
hispanic\_latino & -0.2547*** & No vs Yes & -0.255*** \\
income & 0.1481*** & Less than \$10,000 vs More than \$200,000 & 0.386*** \\
education & -0.0216*** & Bachelor's degree vs Some college or associate's degree & -0.306*** \\
employment\_status & 0.0702*** & On disability vs Unemployed & 0.422*** \\
living\_situation & 0.0610*** & I live alone vs I live with a spouse or romantic partner & -0.390*** \\
political\_id & 0.0365*** & Moderate vs Very conservative & -0.275*** \\
gender & -0.6329*** & Female vs Male & -0.633*** \\
\bottomrule
\end{tabular}%
}
\caption{Summary of Spearman Correlations by Demographic Category for \textbf{Sentiment Analysis dataset}. {\small \textit{Note:} *** $p<.001$, ** $p<.01$, * $p<.05$, ns = not significant.}}
\label{tab:spearman_summary_sentiment}
\end{table*}

\begin{table*}[htbp]
\centering
\small
\resizebox{\textwidth}{!}{%
\begin{tabular}{lrlr}
\toprule
\textbf{Demographic} & \textbf{Fisher $\bar{\rho}$} & \textbf{Dominant Pair} & \textbf{Dominant $\rho$} \\
\midrule
race & 0.1440*** & Black or African American vs White & -0.502*** \\
gender & -0.0462*** & Man vs Woman & -0.459*** \\
age & 0.0483*** & 50-54 vs 54-59 & 0.243*** \\
education & 0.0275*** & College degree vs High school diploma or equivalent & -0.429*** \\
occupation & 0.0534*** & Prefer not to disclose vs Student & 0.474*** \\
\bottomrule
\end{tabular}%
}
\caption{Summary of Spearman Correlations by Demographic Category for \textbf{Politeness dataset}. {\small \textit{Note:} *** $p<.001$, ** $p<.01$, * $p<.05$, ns = not significant.}}
\label{tab:spearman_summary_politeness}
\end{table*}

\begin{table*}[htbp]
\centering
\small
\resizebox{\textwidth}{!}{%
\begin{tabular}{lrlr}
\toprule
\textbf{Demographic} & \textbf{Fisher $\bar{\rho}$} & \textbf{Dominant Pair} & \textbf{Dominant $\rho$} \\
\midrule
race & 0.0580*** & Arab American vs Native American & 0.439*** \\
gender & -0.1823*** & Man vs Woman & -0.319*** \\
age & 0.1385*** & 25-29 vs 60-64 & 0.305*** \\
education & -0.0534*** & Less than a high school diploma vs Other & 0.212** \\
occupation & 0.0967*** & Other vs Prefer not to disclose & 0.417*** \\
\bottomrule
\end{tabular}%
}
\caption{Summary of Spearman Correlations by Demographic Category for \textbf{Offensiveness dataset}. {\small \textit{Note:} *** $p<.001$, ** $p<.01$, * $p<.05$, ns = not significant.}}
\label{tab:spearman_summary_offensiveness}
\end{table*}

\begin{table*}[htbp]
\centering
\small
\resizebox{\textwidth}{!}{%
\begin{tabular}{lrlr}
\toprule
\textbf{Demographic} & \textbf{Fisher $\bar{\rho}$} & \textbf{Dominant Pair} & \textbf{Dominant $\rho$} \\
\midrule
AGE & 0.2804*** & 28.0 vs 31.0 & 0.506*** \\
sex\_map & -0.0874*** & female vs male & -0.327*** \\
reli\_map & 0.2178*** & native vs others & 0.544*** \\
focc\_map & 0.2647*** & Unknown vs unemployed & 0.511*** \\
mocc\_map & 0.2647*** & Unknown vs self-employed nonacademic & 0.444*** \\
fiel\_map & 0.3071*** & natural science vs other & 0.517*** \\
coun\_map & 0.3557*** & australia vs usa & 0.548*** \\
\bottomrule
\end{tabular}%
}
\caption{Summary of Spearman Correlations by Demographic Category for \textbf{Emotion dataset}. {\small \textit{Note:} *** $p<.001$, ** $p<.01$, * $p<.05$, ns = not significant.}}
\label{tab:spearman_summary_emotion}
\end{table*}

\section{ Per-Task Analysis of RQ2}
\label{app:rq2_detail}

Below we provide per-task commentary connecting the three dataset-side analyses
(SHAP importance from Table~\ref{tab:shap_all_tasks}, LSVC learnability from Table~\ref{tab:lsvc_performance_all}, and Fisher-averaged directional coherence from Tables~\ref{tab:spearman_summary_toxicity}--\ref{tab:spearman_summary_emotion})
to the LLM alignment outcomes.

\paragraph{Toxicity.}
Toxicity provides the strongest evidence for dataset-side signal predicting alignment.
SHAP ranks \texttt{religion\_important} highest (0.189), yet its Fisher $\bar{\rho}$ is near zero ($-0.001$), indicating that its subgroups do not agree on which words indicate toxicity.
By contrast, \texttt{race} (SHAP: $0.106$, rank 6/8) and \texttt{lgbtq\_status} (SHAP: $0.100$, rank 7/8) carry the highest positive Fisher $\bar{\rho}$ ($+0.250$ and $+0.208$) and also rank among the top three LSVC $\kappa$ values ($0.186$ and $0.200$, Table~\ref{tab:lsvc_performance_all}).
These are precisely the attributes anchoring the best-performing multi-component configurations in Table~\ref{tab:combined_toxicity_kappa_components}: \texttt{lgbtq\_status} appears in every top-5 configuration for Llama, Qwen, and Mistral.
The LSVC Spearman correlation is positive for Gemma ($0.524$) and significant for Qwen ($\mathbf{0.714}$, $p < 0.05$).
This task illustrates the key principle: an attribute can be a strong predictor of human label variation (high SHAP) without offering a coherent signal that a prompted LLM can exploit; conversely, attributes with moderate SHAP importance but high directional coherence enable the strongest alignment gains.

\paragraph{Sentiment.}
All five LSVC Spearman correlations are positive ($0.176$--$0.502$), but none reaches significance.
The Fisher $\bar{\rho}$ analysis explains the limited alignment gains on this task.
\texttt{Gender}---one of the few attributes occasionally beating baseline (Llama: $\kappa = 0.350$ vs.\ $0.347$; Qwen: $0.349$ vs.\ $0.337$)---carries the most severely opposed signal in the entire study ($\bar{\rho} = -0.633$).
Even attributes with positive Fisher $\bar{\rho}$ (e.g., \texttt{race}: $+0.207$; \texttt{income}: $+0.148$) produce only marginal alignment improvements, and the ``all together'' configuration degrades every model substantially (Table~\ref{tab:combined_sentiment_kappa_components}).
The positive but non-significant LSVC correlations suggest that learnability and alignment point in the same direction on this task, but the overall weakness of directionally coherent signal limits the practical utility of demographic prompting for Sentiment.

\paragraph{Politeness.}
Politeness exhibits a split: Gemma ($0.400$) and Qwen ($0.700$) show positive LSVC correlations, while DeepSeek ($-0.500$) is negative.
Gemma is the sole model benefiting from demographic prompting, peaking with \texttt{occupation} ($\kappa = 0.411$) and \texttt{gender} ($0.408$).
\texttt{Occupation} has the fourth-highest LSVC $\kappa$ on Politeness ($0.472$, Table~\ref{tab:lsvc_performance_all}) and a modestly positive Fisher $\bar{\rho}$ ($+0.053$) with a strong dominant-pair correlation of $+0.474$.
By contrast, \texttt{gender}---the highest-LSVC attribute ($0.507$)---carries a negative Fisher $\bar{\rho}$ ($-0.046$; Man vs.\ Woman $\rho = -0.459$), meaning its high LSVC learnability reflects classifier capacity to exploit the \emph{divergence} between subgroups rather than a coherent shared signal.
That Gemma nonetheless benefits from \texttt{gender} prompting suggests model-specific factors beyond dataset-side signal also play a role.

\paragraph{Offensiveness.}
The systematic negative LSVC correlations ($\rho \leq -0.700$ for four of five models) are the most striking result in Table~\ref{tab:lsvc_spearman_results}.
The LSVC ranks \texttt{age} highest ($\kappa = 0.316$), followed by \texttt{education} ($0.310$) and \texttt{gender} ($0.306$), yet these are among the attributes that most degrade LLM alignment (e.g., \texttt{age} drops Gemma from $0.248$ to $0.152$).
Meanwhile, \texttt{race}---ranked lowest by LSVC ($0.302$)---is the only attribute that significantly improves any model (Qwen: $0.256$ vs.\ $0.205$).
The Fisher analysis clarifies: \texttt{gender} ($\bar{\rho} = -0.182$) is directionally opposed, and \texttt{age} ($\bar{\rho} = +0.139$) has moderate coherence that does not translate to persona-exploitable structure.
\texttt{Race} ($\bar{\rho} = +0.058$) has low overall coherence but may contain specific subgroup pairings that Qwen can exploit.
DeepSeek is the exception ($\rho = +0.700$), though this does not correspond to actual alignment gains---DeepSeek's baseline remains its best performance on every task.

\paragraph{Emotion.}
The dominant alignment attribute (\texttt{age}, which lifts Mistral from $0.483$ to $0.611$ and Qwen from $0.603$ to $0.641$) ranks only fourth in LSVC accuracy ($0.539$, Table~\ref{tab:lsvc_performance_all}), behind \texttt{country} ($0.577$), \texttt{field\_of\_study} ($0.559$), and \texttt{religion} ($0.546$).
Despite their higher learnability, these top-LSVC attributes produce worse alignment: \texttt{country} drops Gemma from $0.726$ to $0.474$; \texttt{field\_of\_study} drops it to $0.418$.
Fisher coherence does not resolve this: \texttt{country} ($\bar{\rho} = +0.356$) and \texttt{field\_of\_study} ($\bar{\rho} = +0.307$) both exceed \texttt{age} ($\bar{\rho} = +0.280$).
This discrepancy reinforces the conclusion from the main text: dataset-side signal---even when both learnable and directionally coherent---does not deterministically predict which attributes a specific model architecture can successfully leverage.
The negative LSVC correlations for Gemma ($-0.523$), Llama ($-0.667$), and Mistral ($-0.541$) reflect the fact that \texttt{age} drives alignment for these models despite not ranking highest on either LSVC learnability or Fisher coherence.

\section{Detailed Results for Neuron Probing}
\label{app:neuron_probing}

\subsection{Ablation Study}
\label{app:ablation}

\begin{figure*}[t]
\centering
\includegraphics[width=1.0\linewidth]{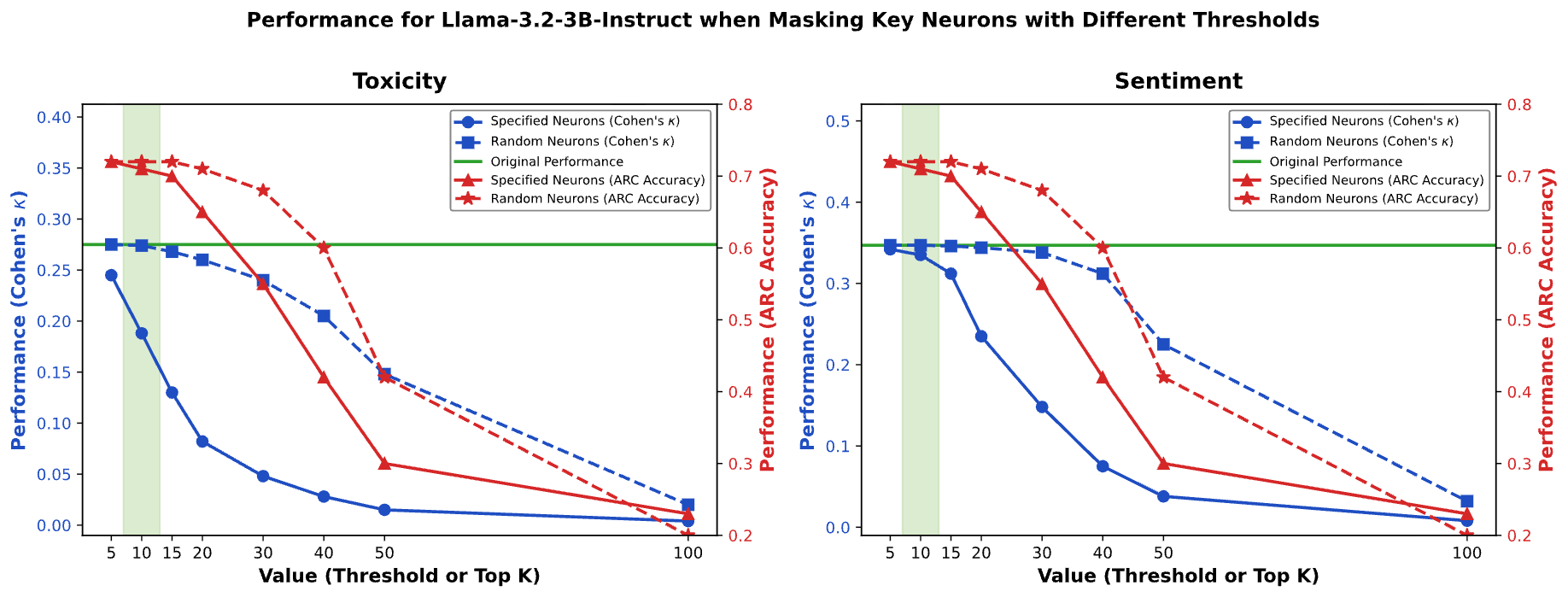}
\caption{Ablation study for Toxicity and Sentiment Task}
\label{fig:ablation_study}
\end{figure*}

To determine the optimal threshold for Key Neuron selection, we evaluate the effect of masking neurons at varying top-$k$ values ($k \in \ {5, 10, 15, 20, 30, 40, 50, 100}$) on both in-task performance (Cohen's $\kappa$) and out-of-distribution performance (ARC accuracy \cite{clark2018think}). \Cref{fig:ablation_study} shows results for Llama-3.2-3B-Instruct on Toxicity and Sentiment. At each threshold, we compare masking the specified top-k neurons (selected by activation magnitude) against masking the same number of randomly chosen neurons. At $k= 10$, masking specified neurons produces a significant drop in task Cohen $\kappa$ while ARC accuracy remains largely unaffected, confirming that these neurons encode task-specific rather than general-purpose knowledge. Random masking at the same scale causes minimal degradation on either metric, verifying that the selected neurons are not arbitrary. At higher thresholds ($k \ge 30$), both specified and random masking degrade ARC performance substantially, indicating that the masking volume exceeds the task-specific neuron population and begins destroying general model capacity. We therefore set $k$ = 10 for all subsequent experiments, as it captures task-relevant neurons while preserving the model's broader representational integrity.

\subsection{Selective Support on Toxicity}

The hypothesis for RQ3 receives its clearest support on Toxicity.
Table~\ref{tab:neuron_count_correlation} shows statistically significant
positive Pearson correlations for Llama ($r = 0.742$) and Qwen
($r = 0.769$): attributes that activate a larger proportion of new/specialized
neurons tend to produce better alignment with human toxicity
annotations.
The per-attribute data (Table~\ref{tab:specialized_toxicity})
make the pattern concrete: for Qwen, \texttt{lgbtq\_status} and
\texttt{religion\_important} activate the highest specialised-neuron proportions
($0.436$ and $0.387$) and also yield the strongest single-component
alignment gains ($\kappa = 0.246$ and $0.242$ vs.\ baseline $0.157$, \Cref{tab:combined_toxicity_kappa_components}).
For Mistral, \texttt{lgbtq\_status} similarly activates the most
specialised neurons ($0.422$) and achieves the highest single-component
$\kappa$ ($0.316$).
Crucially, these are precisely the attributes identified in
Section~\ref{sec:directional_coherence} as carrying directionally coherent lexical
signals (Fisher $\bar{\rho} = +0.208$ for \texttt{lgbtq\_status} and
$+0.250$ for \texttt{race}, \Cref{tab:spearman_summary_toxicity}), suggesting that specialised neuron
activation is a meaningful correlate of alignment only when the
underlying annotation signal is structurally exploitable.

\subsection{The DeepSeek High-Volume Paradox}

A striking counter-example emerges from DeepSeek.
Across all five tasks, DeepSeek activates substantially more specialised
neurons than any other model (see
Tables~\ref{tab:specialized_toxicity}--\ref{tab:specialized_emotion},
Appendix).
If the specialised-neuron hypothesis held universally, DeepSeek should
be the most demographically steerable model.
The alignment results show precisely the opposite: DeepSeek is the only
model whose no-demographic baseline is its best configuration on every
task, and no single-component prompt produces a statistically
significant improvement over baseline on any task.

We term this the \textit{high-volume paradox}.
One interpretation, consistent with the mechanistic interpretability
literature~\citep{cao2025model}, is that activation quantity does not
index the quality or relevance of knowledge engaged: DeepSeek's high
neuron volumes may reflect broad, undifferentiated representational
perturbation in response to persona injection rather than targeted
engagement with annotation-relevant knowledge.
% This is consistent with DeepSeek's small model size (1.5B parameters),
% which may limit its capacity to maintain stable task representations
% under augmented context.

\subsection{Negative Correlations and Task-Dependent Breakdown}

Table~\ref{tab:neuron_count_correlation} reveals that the
neuron--performance correlation is significantly \textit{negative}
for several model--task combinations.
On Sentiment, Gemma ($r = -0.888$) and Mistral ($r = -0.727$) both
show strong, statistically significant negative correlations; on Emotion,
Gemma reaches $r = -0.983$, the largest-magnitude value in the
table and Llama shows $r = -0.772$.
The structural explanation connects directly to the dataset-level
analysis in Section~\ref{sec:directional_coherence}. For Gemma on Sentiment, the attributes activating the \textit{most}
specialised neurons \texttt{political\_id} ($0.288$) and
\texttt{hispanic\_latino} ($0.262$) also produce the largest
alignment degradations: \texttt{political\_id} drops Gemma's $\kappa$
from $0.388$ to $0.300$.
Conversely, \texttt{gender}, which activates the \textit{fewest}
specialised neurons for Gemma on this task ($0.183$), causes the
smallest degradation ($\kappa = 0.366$).
The key factor is signal coherence: \texttt{gender} on Sentiment carries
the most severely opposed annotation signal in the study (Fisher
$\bar{\rho} = -0.633$), meaning additional neuron activation engages
a structurally irreconcilable subgroup divergence that no single persona
can resolve.
The Emotion results follow the same logic: Gemma's strong baseline
($0.726$, the highest on that task) is disrupted by the demographic
attributes that activate the most new neurons \texttt{field\_of\_study}
($0.696$) and \texttt{country} ($0.671$) produce the steepest accuracy
drops, to $0.418$ and $0.474$ respectively.
High specialised-neuron activation here reflects interference with an
already well-calibrated representation, not useful knowledge engagement.

\subsection{Gemma's Politeness Exception}

Gemma is the only model that benefits from demographic prompting on
Politeness, with \texttt{occupation} ($\kappa = 0.411$),
\texttt{gender} ($0.408$), and \texttt{race} ($0.407$) all exceeding
the baseline ($0.372$).
Yet the Pearson correlation for Gemma on Politeness is negative
($r = -0.824$), and Table~\ref{tab:specialized_politeness} (Appendix)
shows that \texttt{occupation} (the best-performing attribute) activates
the \textit{fewest} specialised neurons for Gemma ($0.595$), while
\texttt{age} activates the most ($0.633$).
The resolution again lies in signal quality: \texttt{occupation} carries
the most directionally coherent Fisher-averaged signal on this task
($\bar{\rho} = +0.053$, dominant-pair $\rho = +0.474$), while
\texttt{age} ($\bar{\rho} = +0.048$) and \texttt{race}
($\bar{\rho} = +0.144$ but dominant Black vs.\ White pair
$\rho = -0.502$) carry more structurally conflicted signals.
Specialised neuron count is once more a poor proxy for alignment
utility when signal coherence is not accounted for.

\begin{table}[ht]
\centering
\small
\resizebox{\columnwidth}{!}{%
\begin{tabular}{lccccc}
\toprule
\textbf{Component} & \textbf{Gemma} & \textbf{Llama} & \textbf{Qwen} & \textbf{DeepSeek} & \textbf{Mistral} \\
\midrule
age\_range & 0.4413 & 0.4361 & 0.3647 & 0.7450 & 0.3759 \\
education & 0.4163 & 0.4450 & 0.3737 & 0.7550 & 0.3820 \\
gender & 0.4471 & 0.4385 & 0.3691 & 0.7508 & 0.3725 \\
is\_parent & 0.4270 & 0.4279 & 0.3634 & 0.7820 & 0.3788 \\
lgbtq\_status & 0.4317 & 0.4834 & 0.4360 & 0.7312 & 0.4220 \\
political\_affilation & 0.4273 & 0.4351 & 0.3678 & 0.7003 & 0.3814 \\
race & 0.4384 & 0.4908 & 0.3764 & 0.7166 & 0.3836 \\
religion\_important & 0.4425 & 0.4456 & 0.3879 & 0.7953 & 0.3801 \\
\bottomrule
\end{tabular}%
}
\caption{Specialized Neurons across Components and Models for \textbf{Toxicity}}
\label{tab:specialized_toxicity}
\end{table}

\begin{table}[ht]
\centering
\small
\resizebox{\columnwidth}{!}{%
\begin{tabular}{lccccc}
\toprule
\textbf{Component} & \textbf{Gemma} & \textbf{Llama} & \textbf{Qwen} & \textbf{DeepSeek} & \textbf{Mistral} \\
\midrule
age & 0.2127 & 0.3072 & 0.3075 & 0.7402 & 0.3384 \\
education & 0.2157 & 0.3115 & 0.3006 & 0.7486 & 0.3317 \\
employment\_status & 0.2190 & 0.3171 & 0.3099 & 0.7607 & 0.3413 \\
gender & 0.1827 & 0.3053 & 0.2897 & 0.7211 & 0.3312 \\
hispanic\_latino & 0.2617 & 0.3159 & 0.3100 & 0.6928 & 0.3484 \\
income & 0.2560 & 0.3153 & 0.3202 & 0.7944 & 0.3389 \\
living\_situation & 0.2118 & 0.3142 & 0.3070 & 0.7563 & 0.3396 \\
political\_id & 0.2880 & 0.3358 & 0.3148 & 0.6979 & 0.3528 \\
race & 0.2191 & 0.3266 & 0.3144 & 0.7414 & 0.3486 \\
\bottomrule
\end{tabular}%
}
\caption{Specialized Neurons across Components and Models for \textbf{Sentiment}}
\label{tab:specialized_sentiment}
\end{table}

\begin{table}[ht]
\centering
\small
\resizebox{\columnwidth}{!}{%
\begin{tabular}{lccccc}
\toprule
\textbf{Component} & \textbf{Gemma} & \textbf{Llama} & \textbf{Qwen} & \textbf{DeepSeek} & \textbf{Mistral} \\
\midrule
age & 0.6328 & 0.5247 & 0.4606 & 0.6955 & 0.4504 \\
education & 0.6243 & 0.5169 & 0.4506 & 0.6946 & 0.4476 \\
gender & 0.6079 & 0.5327 & 0.4562 & 0.6993 & 0.4498 \\
occupation & 0.5952 & 0.5108 & 0.4439 & 0.6916 & 0.4558 \\
race & 0.6216 & 0.5529 & 0.4524 & 0.6964 & 0.4500 \\
\bottomrule
\end{tabular}%
}
\caption{Specialized Neurons across Components and Models for \textbf{Politeness}}
\label{tab:specialized_politeness}
\end{table}

\begin{table}[ht]
\centering
\small
\resizebox{\columnwidth}{!}{%
\begin{tabular}{lccccc}
\toprule
\textbf{Component} & \textbf{Gemma} & \textbf{Llama} & \textbf{Qwen} & \textbf{DeepSeek} & \textbf{Mistral} \\
\midrule
age & 0.4722 & 0.4131 & 0.3765 & 0.7188 & 0.4242 \\
education & 0.4468 & 0.4044 & 0.3849 & 0.7175 & 0.4183 \\
gender & 0.4546 & 0.4108 & 0.3837 & 0.7232 & 0.4216 \\
occupation & 0.4458 & 0.3967 & 0.3706 & 0.7048 & 0.4315 \\
race & 0.4723 & 0.4442 & 0.3844 & 0.7058 & 0.4333 \\
\bottomrule
\end{tabular}%
}
\caption{Specialized Neurons across Components and Models for \textbf{Offensiveness}}
\label{tab:specialized_offensiveness}
\end{table}

\begin{table}[ht]
\centering
\small
\resizebox{\columnwidth}{!}{%
\begin{tabular}{lccccc}
\toprule
\textbf{Component} & \textbf{Gemma} & \textbf{Llama} & \textbf{Qwen} & \textbf{DeepSeek} & \textbf{Mistral} \\
\midrule
age & 0.5734 & 0.5384 & 0.3666 & 0.7031 & 0.4510 \\
country & 0.6714 & 0.5800 & 0.3382 & 0.7092 & 0.4585 \\
father\_occupation & 0.6665 & 0.5577 & 0.3831 & 0.6953 & 0.4613 \\
field\_of\_study & 0.6959 & 0.5652 & 0.3596 & 0.6888 & 0.4580 \\
gender & 0.5828 & 0.5557 & 0.3638 & 0.6837 & 0.4524 \\
mother\_occupation & 0.6251 & 0.5366 & 0.3661 & 0.6996 & 0.4544 \\
religion & 0.6163 & 0.5664 & 0.3578 & 0.7024 & 0.4612 \\
\bottomrule
\end{tabular}%
}
\caption{Specialized Neurons across Components and Models for \textbf{Emotion}}
\label{tab:specialized_emotion}
\end{table}

\section{Refusal}
\label{app:refusal}

\begin{table*}[h]
    \centering
    {\small
    \begin{tabular}{lrrrrr}
        \toprule
        \textbf{Model} & \textbf{Toxicity} & \textbf{Sentiment} & \textbf{Politeness} & \textbf{Offensiveness} & \textbf{Emotion} \\
        \midrule
        Gemma3-12B     & 1,204 (1.12\%) & 142 (0.58\%) & 387 (1.55\%) & 892 (0.83\%) & 98 (0.41\%) \\
        Llama-3.2-3B   & 4,312 (4.01\%) & 876 (3.61\%) & 621 (2.48\%) & 2,147 (3.29\%) & 412 (1.72\%) \\
        Qwen2.5-7B     & 83 (0.08\%) & 12 (0.05\%) & 27 (0.11\%) & 54 (0.08\%) & 8 (0.03\%) \\
        DeepSeek-R1-7B & 1,847 (1.72\%) & 324 (1.34\%) & 498 (1.99\%) & 1,103 (1.69\%) & 215 (0.90\%) \\
        Mistral-7B     & 956 (0.89\%) & 187 (0.77\%) & 312 (1.25\%) & 645 (0.99\%) & 104 (0.43\%) \\
        \bottomrule
    \end{tabular}
    }
        \caption{Task-wise refusal rates for LLMs, averaged across all prompting templates.}
    \label{tab:task_refusals}\vspace{-2mm}
\end{table*}

Table~\ref{tab:task_refusals} reports task-wise refusal rates averaged across all 3 prompting templates. Refusal patterns vary substantially across models and tasks. Llama-3.2-3B exhibits the highest refusal rates across all five tasks, peaking at 4.01\% on Toxicity, likely due to its safety-tuning behavior/conservative guardrails at smaller model scale. In contrast, Qwen2.5-7B produces the fewest refusals, remaining below 0.11\% on every task, suggesting more permissive output behavior under persona prompting. Across tasks, Toxicity and Offensiveness elicit the most refusals, consistent with the sensitive nature of these classification targets models are more likely to decline rating content that contains potentially harmful language. Sentiment and Emotion produce the lowest refusal rates, reflecting the comparatively neutral framing of these tasks. These refusals were excluded from all alignment calculations reported in the main text; including them as incorrect predictions does not change the relative ordering of model performance.

\section{Practical Suggestions for Demographic Prompting}
\label{sec:practical-suggestions}
 
Our findings reveal that demographic prompting is not a one-size-fits-all strategy: its effectiveness depends on the interaction between model architecture, task characteristics, and the structural quality of the demographic signal being prompted. Below, we distill actionable guidance organized by task and model, grounded in the empirical patterns from Sections~\ref{sec:rq1_dis} through~\ref{sec:rq3_dis}.
 
\paragraph{Toxicity detection is the most favorable setting for demographic prompting, but attribute selection is critical.}
Across four of five models, compact demographic prompts (one to three attributes) improve alignment over the no-demographic baseline on toxicity (Table~\ref{tab:combined_toxicity_kappa_components}). Users seeking to improve LLM--human alignment on toxicity tasks should prioritize attributes with high directional coherence, such as \texttt{lgbtq\_status} ($\bar{\rho} = +0.208$) and \texttt{race} ($\bar{\rho} = +0.250$), which anchor the top-performing configurations for Llama, Qwen, and Mistral. Conversely, attributes with high SHAP importance but low coherence, such as \texttt{religion\_important} (Fisher $\bar{\rho} \approx 0$), should be used cautiously despite their strong influence on human label variation. Users should avoid prompting with the full attribute set, as the 8-component configuration never outperforms the best compact prompt and often degrades performance significantly.
 
\paragraph{For sentiment analysis, demographic prompting offers limited and model-specific benefits; a conservative approach is advisable.}
Sentiment presents weak and inconsistent gains from demographic prompting (Figure~\ref{fig:example_sentiment}). For Gemma and Mistral, no demographic prompt surpasses baseline, so users of these models should default to unprompted classification. Qwen users may see modest improvement with \texttt{education} + \texttt{hispanic\_latino}, and Llama users with \texttt{gender} alone, but these gains are small. The severely opposed signal carried by \texttt{gender} on this task ($\bar{\rho} = -0.633$) means that gender-based persona prompts are structurally unlikely to help, despite the intuitive appeal of gendered perspectives on sentiment. Users working on sentiment tasks should generally avoid multi-attribute prompts, as the ``all together'' configuration degrades every model substantially.
 
\paragraph{Politeness and offensiveness tasks are broadly resistant to demographic prompting; users should prefer unprompted baselines unless model-specific exceptions apply.}
On politeness, four of five models perform best at baseline, with demographic attributes often causing significant degradation (e.g., \texttt{race} drops Llama's $\kappa$ from 0.349 to 0.252). The sole exception is Gemma, where single-attribute prompts such as \texttt{occupation} and \texttt{gender} improve alignment. Users of Gemma on politeness tasks may therefore experiment with these attributes individually, but should avoid combining them, as multi-attribute configurations erode the gains. On offensiveness, \texttt{race} is the only attribute that selectively improves alignment (for Qwen and Mistral), while \texttt{age} and \texttt{gender}, despite being the most learnable attributes by LSVC $\kappa$, consistently degrade performance due to their opposed subgroup signals. Users should treat high learnability as a necessary but insufficient condition: an attribute that a classifier can exploit may still be unusable by a persona prompt if its subgroups disagree directionally.
 
\paragraph{On emotion attribution, \texttt{age} is the dominant beneficial attribute, but only for select models.}
Mistral and Qwen benefit substantially from \texttt{age}-anchored prompts on the emotion task (Mistral improves by 12.9 percentage points; Qwen by 3.8 pp), making \texttt{age} the recommended single attribute for users of these models on emotion-related tasks. Llama users may achieve smaller gains through specific 4- and 5-way combinations anchored on \texttt{age}, \texttt{gender}, and parental occupations, though the improvement over baseline is marginal. Users of Gemma and DeepSeek should avoid demographic prompting entirely on this task, as every demographic configuration degrades their strong baselines without exception.
 
\paragraph{DeepSeek users should not use demographic prompting under current conditions.}
DeepSeek is the only model whose no-demographic baseline represents its best configuration on every task examined. No single-attribute or multi-attribute prompt produces a statistically significant improvement. The high-volume paradox identified in our neuron probing analysis (Section~\ref{sec:rq3_dis}) suggests that DeepSeek's broad but undifferentiated neuron activation in response to persona cues reflects representational perturbation rather than targeted engagement with annotation-relevant knowledge. This may be a consequence of the distillation process (see Limitations), but regardless of the cause, the practical recommendation is clear: users of DeepSeek-R1-Distill-Qwen-7B should rely on unprompted inference for all five tasks.
 
\paragraph{When in doubt, prefer fewer attributes and verify signal quality before prompting.}
The over-specification threshold documented in RQ1, where alignment consistently degrades beyond one to three attributes, provides a general heuristic: \textit{less is more} in demographic prompting. Users who must decide which attributes to include should, where feasible, assess (1) whether the attribute's subgroups exhibit directionally coherent annotation signals (positive Fisher $\bar{\rho}$), and (2) whether the attribute contributes learnable lexical signal (high LSVC $\kappa$). Attributes satisfying both conditions (e.g., \texttt{lgbtq\_status} and \texttt{race} on toxicity) are the strongest candidates. Attributes with high learnability but opposed signals (e.g., \texttt{gender} on sentiment or offensiveness) should be excluded, as they create structurally irreconcilable demands that no single persona prompt can satisfy. In the absence of dataset-level diagnostics, defaulting to the unprompted baseline is the safest strategy.

\end{document}